\documentclass{article} 
\usepackage[accepted]{icml2026}

\usepackage[utf8]{inputenc} 
\usepackage[T1]{fontenc}    
\usepackage{hyperref}       
\usepackage{url}            
\usepackage{booktabs}       
\usepackage{amsfonts}       
\usepackage{nicefrac}       
\usepackage{microtype}      
\usepackage{xcolor}         
\usepackage{xfrac}
\usepackage{amsmath,amsthm,amssymb}
\usepackage{pythonhighlight}
\usepackage{graphicx}
\usepackage[export]{adjustbox}
\usepackage{longtable}
\usepackage{booktabs}
\usepackage{tabularx}

\icmltitlerunning{APEX: Approximate-but-Exhaustive Search for Ultra-Large Combinatorial Synthesis Libraries}

\newcommand{\appropto}{\mathrel{\vcenter{
  \offinterlineskip\halign{\hfil$##$\cr
    \propto\cr\noalign{\kern2pt}\sim\cr\noalign{\kern-2pt}}}}}

\begin{document}

\twocolumn[
  \icmltitle{APEX: Approximate-but-Exhaustive Search for Ultra-Large Combinatorial Synthesis Libraries}



  \icmlsetsymbol{equal}{*}

  \begin{icmlauthorlist}
    \icmlauthor{Aryan Pedawi}{equal,numerion}
    \icmlauthor{Jordi Silvestre-Ryan}{equal,numerion}
    \icmlauthor{Bradley Worley}{numerion}
    \icmlauthor{Darren J. Hsu}{nvidia}
    \icmlauthor{Kushal S. Shah}{nvidia}
    \icmlauthor{Elias Stehle}{nvidia}
    \icmlauthor{Jingrong Zhang}{nvidia}
    \icmlauthor{Izhar Wallach}{numerion}
  \end{icmlauthorlist}

  \icmlaffiliation{numerion}{Work performed at \mbox{Numerion Labs}}
  \icmlaffiliation{nvidia}{NVIDIA}

  \icmlcorrespondingauthor{Aryan Pedawi}{pedawia@gmail.com}
  \icmlcorrespondingauthor{Jordi Silvestre-Ryan}{jordisr@berkeley.edu}


  \vskip 0.3in
]



\printAffiliationsAndNotice{\icmlEqualContribution}




%


\begin{abstract}
Make-on-demand combinatorial synthesis libraries (CSLs) like Enamine REAL have significantly enabled drug discovery efforts. However, their large size presents a challenge for virtual screening, where the goal is to identify the top compounds in a library according to a computational objective (e.g., optimizing docking score) subject to computational constraints under a limited computational budget. For current library sizes---numbering in the tens of billions of compounds---and scoring functions of interest, a routine virtual screening campaign may be limited to scoring fewer than 0.1\% of the available compounds, leaving potentially many high scoring compounds undiscovered. Furthermore, as constraints (and sometimes objectives) change during the course of a virtual screening campaign, existing virtual screening algorithms typically offer little room for amortization. We propose the approximate-but-exhaustive search protocol for CSLs, or APEX. APEX utilizes a neural network surrogate that exploits the structure of CSLs in the prediction of objectives and constraints to make full enumeration on a consumer GPU possible in under a minute, allowing for exact retrieval of approximate top-$k$ sets. To demonstrate APEX's capabilities, we develop a benchmark CSL comprised of more than 10 million compounds, all of which have been annotated with their docking scores on five medically relevant targets along with physicohemical properties measured with RDKit such that, for any objective and set of constraints, the ground truth top-$k$ compounds can be identified and compared against the retrievals from any virtual screening algorithm. We show APEX's consistently strong performance both in retrieval accuracy and runtime compared to alternative methods.\footnote{Code is available at \url{https://github.com/NumerionLabs/apex}. Please refer to the README for links to data and model weights.}
\end{abstract}

\section{Introduction}

The search for novel therapeutic agents is a cornerstone of modern medicine and drug discovery. In recent years, the emergence of ultra-large combinatorial synthesis libraries (CSLs), such as the Enamine REAL library, has significantly transformed this pursuit. These libraries, containing billions or even trillions of make-on-demand compounds, offer an unprecedented opportunity to explore a vast and diverse chemical space, significantly increasing the potential for identifying novel hit compounds with desirable properties. However, the sheer scale of these libraries presents a formidable challenge to traditional approaches to virtual screening.

State-of-the-art scoring functions used in virtual screening, like docking/affinity/co-folding scores, are too computationally expensive to render an exhaustive evaluation over modern CSLs, which number in the billions, practical. A number of virtual screening approaches have been developed to identify high-scoring compounds from large compound libraries under a limited evaluation budget. These methods include heuristic algorithms \citep{sadybekov2022synthon}, reinforcement learning \citep{pedawi2023through,klarich2024thompson,de2024ngt}, active learning \citep{graff2021accelerating,mehta2021memes}, and approaches that utilize generative models constrained to the library \citep{pedawi2022efficient,cretu2024synflownet,luo2024projecting,gao2025generative}. However, since these algorithms effectively assess only a small fraction of the total library---usually less than 1\% of available compounds---they leave the vast majority of the chemical space unexplored and potentially overlook valuable compounds. Many of the listed strategies above include a surrogate modeling component, in which a more inference efficient model (such as a neural network) is trained to approximate the oracle scoring function to enable exhaustive evaluation with the surrogate \citep{gentile2020deep,graff2022self}. But this too is impeded by the size of modern CSLs, which would naively require billions of neural network evaluations to score exhaustively with the trained surrogate. Indeed, the growing size of CSLs and the computational demands of modern scoring functions in virtual screening create a pressing need for more efficient and comprehensive approaches.

At its core, virtual screening can be framed as a search problem, where the objective is to identify the top $k$ compounds that optimize a specific scoring function while satisfying a set of program-specific constraints, namely desired physicochemical or ADMET properties like molecular weight, lipophilicity, and permeability. The ability to effectively handle constraints is particularly crucial in a virtual screening: for any given drug discovery project, the number of compounds in a screening library that violate these constraints can be orders of magnitude larger than those that satisfy them. This often complicates the workflow and can lead to the exploration of irrelevant chemical space or aggressive post-filtering.

In this work, we introduce APEX (approximate-but-exhaustive search), a new paradigm for searching ultra-large CSLs that enables fast, declarative queries. Once trained, an APEX model allows for efficient retrieval of the (approximate) top-$k$ compounds from a CSL according to a user-specified objective subject to a set of user-specified constraints. More than a virtual screening algorithm, APEX allows for low latency exploration of massive CSLs without the need for a complex, iterative workflow. The core of this capability is a neural network surrogate model that exploits the library's combinatorial structure and amortizes the computation required for repeated querying, enabling real-time search across the entire enumerated CSL with remarkable efficiency on a modern GPU.

This paper details the theoretical underpinnings of the APEX methodology and demonstrates its practical application in virtual screening.

\textbf{Conflict of Interest Disclosure.} At the time this work was conducted, the authors AP, JSR, BR, and IW were employed by Numerion Labs, which develops and utilizes APEX as part of its virtual screening platform. The authors DH, KS, ES, and JZ are employed by NVIDIA, which develops CUDA, in which the factorized top-$k$ is implemented, as well as the NVIDIA GPUs that were utilized in experiments.

\begin{figure*}[ht]
  \centering
  \includegraphics[width=\textwidth]{imgs/apex.pdf}
  \caption{The APEX (approximate-but-exhaustive) search protocol, enabling rapid, on-the-fly virtual screening of ultra-large CSLs. APEX consists of three main steps. \textbf{Step 1: Train the surrogate.} Given an enumerated and labeled dataset, a multi-task neural network is trained to predict molecular properties of interest, like docking scores. \textbf{Step 2: Train the factorizer.} Given a CSL, the reaction factorizer is trained to reconstruct embeddings of the surrogate model from reaction and R-group assignment pairs. The factorizer induces an approximation of surrogate properties that is amenable to substantial amortization in executing top-$k$ retrieval on ultra-large CSLs with respect to those properties. \textbf{Step 3: Run approximate-but-exhaustive search.} Given a search query (e.g., minimize docking score on target of interest subject to drug-likeness constraints), factorized surrogate properties are calculated for all compounds in the CSL and the top-$k$ are retrieved based on the objective subject to constraints. An efficient GPU implementation allows for running a top-$k$ search with $k=\text{1 million}$ on a 10 billion compound CSL in approximately 30 seconds with a single T4 GPU.}
  \label{fig:apex}
\end{figure*}

\section{Related Work}

\textbf{Virtual screening with surrogate models.}
The use of neural network surrogates to approximate expensive scoring functions has been widely explored as a strategy for accelerating virtual screening over large compound libraries. Deep Docking \citep{gentile2020deep} trains iterative QSAR models on a docked subset to progressively filter a library, reducing docking costs to a small fraction of the total. MolPal \citep{graff2021accelerating} frames virtual screening as a Bayesian optimization problem, using a surrogate trained on a growing labeled subset to guide acquisition over a fixed molecular pool; \citet{graff2022self} later extended this with active design-space pruning. MEMES \citep{mehta2021memes} combines Bayesian optimization with deep continuous molecular descriptors to achieve similar gains. These methods are designed for libraries of up to $\sim$$10^8$ molecules and scale linearly with library size during surrogate inference, making exhaustive evaluation with the surrogate a bottleneck as libraries grow into the billions. By exploiting the combinatorial structure of CSLs, APEX learns to decompose surrogate predictions into sums of pre-cached, per-synthon scalar contributions, reducing the cost of scoring any compound in the library to a small number of floating point operations, enabling exact retrieval of approximate top-$k$ sets via a single GPU pass over the full library in seconds.

\textbf{Protein-ligand representation learning for virtual screening.}
A complementary line of work learns joint protein-ligand embeddings for virtual screening as a dense retrieval task. ConPLex \citep{singh2023conplex} co-embeds proteins and drug molecules in a shared feature space using a pretrained protein language model and contrastive learning against decoy compounds, enabling rapid shortlisting across large libraries. DrugCLIP \citep{gao2023drugclip, gao2026drugclip} extends this idea by training on large-scale structural data to align pocket and ligand representations, enabling genome-scale screening at up to ten million times the speed of traditional docking. LigUnity \citep{feng2025ligunity} further unifies virtual screening and hit-to-lead optimization in a single foundation model through a hierarchical contrastive training objective. All of these methods require encoding library compounds individually; as \citet{gao2023drugclip} notes, pre-encoding a 6-billion compound library takes on the order of 30 hours per target, with encoding time scaling linearly with library size. APEX avoids this bottleneck entirely by factorizing predictions over synthon contributions, and is complementary to these methods, as the factorization strategy could in principle be applied on top of such models to enable exhaustive screening against combinatorial synthesis libraries.

\textbf{Synthon-based and hierarchical screening approaches.}
Several methods exploit the building-block structure of CSLs to reduce the effective search space. V-SYNTHES \citep{sadybekov2022synthon} performs hierarchical structure-based screening by first docking minimal enumeration library fragments to identify promising scaffolds and then iteratively elaborating these with synthons to select high-scoring complete molecules, achieving greater than 1000-fold acceleration over exhaustive docking on an 11-billion compound library. Thompson sampling \citep{klarich2024thompson} treats each reaction's synthon pool as a multi-armed bandit and maintains belief distributions over synthon score contributions to guide sequential evaluation. Unlike APEX, both approaches are iterative and require oracle evaluations during search; neither natively handles constraints on molecular properties. APEX, once trained, performs a single exhaustive pass over the full CSL using cached synthon embeddings, handles arbitrary combinations of objectives and constraints without additional oracle calls, and supports rapid amortization across queries.

\textbf{Generative models for synthesizable chemical space.}
An orthogonal approach to large-scale drug discovery uses generative models constrained to synthesizable chemical space. \citet{pedawi2022efficient} introduced a graph generative model, CSLVAE, in which the decoder is constrained to only retrieve compounds from a CSL. NGT \citep{de2024ngt} applies reinforcement learning over a CSLVAE latent space to execute a virtual screen over ultra-large CSLs. Other work, like SynFlowNet \citep{cretu2024synflownet}, use GFlowNets to sample synthesis pathways from building blocks and reaction templates, and methods that project unconstrained generative outputs onto synthesizable analogs \citep{luo2024projecting, gao2025generative}. It should be noted that many of these approaches must effectively retrain or re-optimize the policy for each new screening campaign; changing a constraint or objective can often involve a new round of model fine-tuning or policy optimization, offering little amortization across screens. By contrast, APEX separates the one-time training cost from search execution, allowing modifications of query objectives and constraints to be run without any retraining and quickly.

\textbf{Factorization-based retrieval.}
The use of additive or bilinear factorizations over structured input spaces has a well-established history in machine learning, most notably in recommender systems. Factorization machines \citep{rendle2010fm} and related matrix factorization methods \citep{koren2009matrix} model feature interactions using factorized latent vectors, enabling efficient collaborative filtering at scale. APEX employs an analogous decomposition, i.e., the surrogate embedding of a product molecule is approximated as a sum of per-synthon associative contributions, reducing exhaustive evaluation over tens of billions of compounds to a cached lookup over a few hundred thousand synthon vectors. To our knowledge, APEX is the first method to apply factorization-based retrieval to virtual screening of combinatorial synthesis libraries.

\section{Data}\label{sec:data}

\subsection{Combinatorial Synthesis Libraries}

A combinatorial synthesis library (CSL) is organized into a collection of multi-component \emph{reactions}, each of which has a fixed number of components called \emph{R-groups} which indicate placeholders for molecular building blocks called \emph{synthons}. Hence, each product in a CSL can be identified by its reaction and R-group assignment. Due to their combinatorial design, commercially available make-on-demand CSLs such as the Enamine REAL library span a chemical space numbering in the tens of billions of compounds today from a few hundred thousand synthons. We refer the reader to Appendix \ref{app:csl_structure} for a self-contained description of the CSL formalism and the associated encoder hierarchy.

In this work, we designed our own open CSL as an alternative to existing proprietary CSLs for benchmarking and reproducibility purposes.
We used a random sample of 1 million ``lead-like'' compounds from the ZINC22 database \citep{tingle2023zinc} as a starting point for library construction. 
Our main focus here is on developing a large virtual library of valid CSL-like molecules, so we do not consider or ensure synthetic feasibility.
We used the BRICS fragmentation algorithm \citep{degen2008art}, which breaks specific bonds based on defined chemical environments, to fragment each sampled molecule into two or three fragments.
Each fragment is labeled with numbered pseudoatoms at the break points, with the BRICS rules determining which pseudoatom types can be joined to form a new bond. 
We applied the BRICS rules (as implemented in RDKit) to enumerate two- and three-component reactions that recombined these fragments into valid chemical products. This results in a set of fragmentation rules and fragments analogous to the reactions and synthons of a CSL.

Our final CSL comprises over 10B molecules and is by design evenly split between two- and three-component reactions. Additionally, we generated two smaller libraries by uniformly downsampling each reaction. These smaller libraries contain over 12M and 1M products and are fully enumerated to enable exhaustive docking and calculation of physicochemical properties. To address data leakage concerns, for all experiments in the paper, the surrogate is trained on the 1M compound CSL and evaluation is performed using either the 12M or 10B compound CSLs.

\subsection{Docking Scores and Physicochemical Properties}
For benchmarking purposes, we selected five diverse protein targets: PARP1 (an enzyme), MET (a kinase), DRD2 (a GPCR), F10 (a protease), and ESR1 (a nuclear receptor). Receptor structures and binding sites were obtained from the DOCKSTRING dataset \citep{garcia2022dockstring}.
Molecules from this smaller library were embedded with RDKit and docked against these five targets using an accelerated implementation of the AutoDock Vina \citep{trott2010autodock} scoring function designed to run on GPU \citep{morrison2020cuina}. 
In addition to docking scores, we calculated various physicochemical properties (e.g., molecular weight, number of hydrogen bond donors and acceptors; full list can be seen in Figure \ref{fig:r2}) for each molecule in this enumerated library.

\section{Methodology}

Given a CSL $\mathcal{D}$, which defines the chemical space $X_\mathcal{D}$ of eligible compounds, our goal is to identify the top-$k$ compounds from the library that maximize an objective subject to constraints. This retrieval problem can be expressed as
\begin{align}
\begin{split}
X_k^* \;:=\; \arg\max_{\substack{X_k\subset X_\mathcal{D} \\ |X_k|\le k}} &\quad \sum_{{\bf x}\in X_k}f_0({\bf x}), \\
\text{subject to} &\quad \ell_i\le f_i({\bf x}) \le u_i, \\
& \quad \forall {\bf x}\in X_k, i=1,\ldots,m,
\end{split}\label{orig_topk}
\end{align}
where $f_0: X\to\mathbb{R}$ is the objective and $f_i: X\to\mathbb{R}$, $i=1,\ldots,m$, are the constraints with bounds $\ell_i<u_i$. This problem is complicated by the present and rapidly growing size of CSLs, $|X_\mathcal{D}|>10^{10}$, combined with the fact that many objectives and constraints of interest---such as docking or co-folding scores---are computationally expensive to evaluate, which precludes exhaustive evaluation. We can relax \eqref{orig_topk} by substituting the original objective and constraints with surrogate models:
\begin{align}
\begin{split}
\hat{X}_k^* \;:=\; \arg\max_{\substack{X_k\subset X_\mathcal{D} \\ |X_k|\le k}} &\quad \sum_{{\bf x}\in X_k}\hat{f}_0({\bf x}), \\
\text{subject to} &\quad \ell_i\le \hat{f}_i({\bf x}) \le u_i, \\
& \quad \forall {\bf x}\in X_k, i=1,\ldots,m,
\end{split}\label{approx_topk}
\end{align}
Neural network surrogates that operate directly on a 2D molecular graph or a 1D representation like SMILES are a good choice, but exhaustive evaluation of ultra-large CSLs with such surrogates is still far from a routine computational task, requiring $O(|X_\mathcal{D}|)$ neural network evaluations.

We develop a surrogate-based modeling strategy that permits \eqref{approx_topk} to be solved efficiently for ultra-large CSLs. First, let us discuss the parameterization of the surrogate models admissible under this design.

\subsection{Surrogate Model Parameterization}

Let $g_\theta: X\to\mathbb{R}^d$ be a neural network that encodes a molecule ${\bf x}\in X$ into a $d$-dimensional embedding space. We place no restrictions on $g_\theta$ beyond this, i.e., it can be a transformer that operates on the SMILES representation of ${\bf x}$, a GNN that operates on a 2D graph representation of ${\bf x}$, or some other similarly appropriate choice. We model each task $i=0,\ldots,m$ as a linear function of the molecular embedding,
\begin{align}
    \hat{f}_i({\bf x}) &= {\bf w}_i^\top g_\theta({\bf x}) + b_i,\label{surrogate}
\end{align}
where ${\bf w}_i\in\mathbb{R}^d$ and $b_i\in\mathbb{R}$. Given labeled data from each task, written $p_i({\bf x}, y)$ where $y = f_i({\bf x})$, the surrogate model is trained to minimize the prediction error relative to ground truth:
\begin{align}
    \min_{\theta, \, {\bf W}\in\mathbb{R}^{(m+1)\times d}, \, {\bf b}\in\mathbb{R}^{m+1}} L(\theta, {\bf W}, {\bf b}),
\end{align}
where
\begin{align}
L(\theta, &{\bf W}, {\bf b}) = \nonumber \\
&\sum_{i=0}^m\mathbb{E}_{p_i({\bf x},y)}\mathbb{E}_{p({\pmb\varepsilon})}\left[({\bf w}_i^\top (g_\theta({\bf x}) + {\pmb\varepsilon}) + b_i - y)^2\right].\label{loss_surrogate}
\end{align}
The surrogate is trained with noise added to the embeddings, sampled from a simple distribution $p({\pmb\varepsilon})$ like an isotropic normal. The relevance of this particular detail will be explained shortly.

\subsection{Factorization of Surrogate Embeddings}


As a review (see \citet{pedawi2022efficient} for additional details), we can represent a CSL $\mathcal{D}\equiv (T, R, S, \psi, \phi)$ hierarchically, with \emph{synthons} (molecular fragments) $S$ at the bottom of the hierarchy, \emph{R-groups} $R$ in the middle, and \emph{reactions} $T$ at the top. Every synthon index $s\in S$ is associated with a corresponding molecular representation ${\bf x}_s\in X_*$ (again, SMILES or 2D graph), where $X_*\supset X$ extends $X$ to include attachment points, represented by the token ``*''. An R-group, denoted by the index $r\in R$, is comprised of a set of synthons that constitute valid assignments to the associated component in a multi-component reaction. A multi-component reaction $t\in T$, together with a valid assignment of synthons to their constituent R-groups, produces a single molecule via chemical synthesis as output. We denote by $\psi_{T\to R}: T \to \mathcal{P}(R)$ the function that returns the set of R-groups $\psi_{T\to R}(t) \subset R$ associated with a reaction $t$, where $\mathcal{P}(\cdot)$ denotes the power set function. Similarly, $\psi_{R\to S}: R\to \mathcal{P}(S)$ returns the set of synthons $\psi_{R\to S}(r)\subset S$ that can be assigned to a particular R-group. Each molecule in $\mathcal{D}$ can be referenced by a multi-index, denoted by ${\pmb\chi} = (t, \{(r, s) : \exists s\in\psi_{R\to S}(r), \forall r\in \psi_{T\to R}(t)\})$, which describes the reaction and R-group assignment used to construct the molecule, ${\bf x} := \phi({\pmb\chi})$.

We utilize the design proposed in \citet{pedawi2022efficient} to model an associated hierarchy of representations that describe the library at these three levels of resolution. First, the $\texttt{SynthonEncoder}: X_*\to\mathbb{R}^{d_S}$ produces an embedding for each synthon $s$ as a function of its molecular representation ${\bf x}_s$. Next, a deep set network called the $\texttt{RgroupEncoder}: \mathbb{R}^{d_S}\times\cdots\times\mathbb{R}^{d_S}\to\mathbb{R}^{d_R}$ produces an embedding for each R-group $r$ as a function of the representations of its constituent synthons. Finally, another deep set network, $\texttt{ReactionEncoder}:\mathbb{R}^{d_R}\times\cdots\times\mathbb{R}^{d_R}\to\mathbb{R}^{d_T}$, produces an embedding for each reaction $t$ as a function of the representations of its constituent R-groups. This is described by the following computational stack:
\begin{align}
{\bf h}_s^S &= \texttt{SynthonEncoder}({\bf x}_s), \\
{\bf h}_r^R &= \texttt{RgroupEncoder}(\{{\bf h}_s^S : \forall s\in\psi_{R\to S}(r)\}), \\
{\bf h}_t^T &= \texttt{ReactionEncoder}(\{{\bf h}_r^R : \forall r\in\psi_{T\to R}(t)\}).
\end{align}
From these representations, we aim to reconstruct the molecular embedding $g_\theta(\phi({\pmb\chi}))$ as a function of the associated multi-index ${\pmb\chi}$ in a manner which will permit fast and efficient approximations of \eqref{surrogate}. To do this, we model the embedding space of $g_\theta$ via a linear associative map of the R-group assignments. In particular, we introduce a $\texttt{SynthonValueEncoder}: \mathbb{R}^{d_S}\to\mathbb{R}^{d_U}$ and $\texttt{RgroupKeyEncoder}: \mathbb{R}^{d_R}\times \mathbb{R}^{d_T}\to\mathbb{R}^{d\times d_U}$ which produce intermediate representations that are combined as follows to arrive at a prediction of the molecule's latent representation:
\begin{align}
{\bf v}_s &= \texttt{SynthonValueEncoder}({\bf h}_s^S), \\
{\bf K}_r &= \texttt{RgroupKeyEncoder}({\bf h}_r^R, {\bf h}_{\psi_{R\to T}(r)}^T), \\
{\bf u}_{r,s} &= {\bf K}_r{\bf v}_s, \label{assoc_embeds}\\
\hat{g}_\lambda({\pmb\chi}) &= \sum_{(r,s)\in{\pmb\chi}} {\bf u}_{r,s}. \label{factorizer}
\end{align}
The \texttt{SynthonEncoder}, \texttt{RgroupEncoder}, \texttt{ReactionEncoder}, \texttt{SynthonValueEncoder}, and \texttt{RgroupKeyEncoder} all combine to form the \texttt{ReactionFactorizer} or just the ``factorizer'' for short, which we represent by the function $\hat{g}_{\lambda}({\pmb\chi})$. Given a library $\mathcal{D}$ and the frozen surrogate encoder $g_\theta$, we train the factorizer to minimize the reconstruction error of the surrogate embeddings,
\begin{align}
\min_{\lambda} \quad \mathbb{E}_{p({\pmb\chi}|\mathcal{D})}\left[\|g_\theta(\phi({\pmb\chi})) - \hat{g}_\lambda({\pmb\chi})\|_2^2\right].\label{factorizer_opt}
\end{align}

\subsection{Putting it Together}

We can factorize the surrogate predictions by substituting \eqref{factorizer} into \eqref{surrogate}, which simplifies as follows:
\begin{align}
    \hat{\hat{f}}_i({\pmb\chi}) &= {\bf w}_i^\top\hat{g}_\lambda({\pmb\chi}) + b_i, \label{apex_subtitution}\\
    &= {\bf w}_i^\top\left(\sum_{(r,s)\in{\pmb\chi}} {\bf u}_{r,s}\right) + b_i, \\
    &= \sum_{(r,s)\in{\pmb\chi}}{\bf w}_i^\top {\bf u}_{r,s} + b_i, \\
    &= \sum_{(r,s)\in{\pmb\chi}}v_{i,r,s} + b_i,\label{apex}
\end{align}
where the $v_{i,r,s}$ terms are called \emph{synthon associative contributions}. We use the shorthand $\hat{\hat f}_i({\bf x})$ to denote $\hat{\hat f}_i(\phi({\pmb\chi}))$ when ${\bf x}=\phi({\pmb\chi})$, i.e., we can express $\hat{\hat{f}}_i: X_\mathcal{D}\to\mathbb{R}$.
We call the expression in \eqref{apex} the \emph{approximate-but-exhaustive (APEX) factorization}, because it permits us to solve the top-$k$ problem \eqref{approx_topk} under the approximation \eqref{apex} via exhaustive evaluation on $\mathcal{D}$:
\begin{align}
\begin{split}
\hat{\hat{X}}_k^* \;:=\; \arg\max_{\substack{X_k\subset X_\mathcal{D} \\ |X_k|\le k}} &\quad \sum_{{\bf x}\in X_k}\hat{\hat{f}}_0({\bf x}), \\
\text{subject to} &\quad \ell_i\le 
\hat{\hat{f}}_i({\bf x}) \le u_i, \\
&\quad \forall {\bf x}\in X_k, i=1,\ldots,m.
\end{split}\label{apex_topk}
\end{align}
Since the surrogate is trained with noise added to the embeddings as per \eqref{loss_surrogate} (and therefore learns embeddings whose linear projections are robust to such perturbations), the APEX prediction induced by the substitution in \eqref{apex_subtitution} is robust to the so-called errors-in-variables problem \citep{griliches1974errors}. The addition of isotropic normal noise in \eqref{loss_surrogate} is therefore a technique to statistically regularize the surrogate to ensure that it remains robust to the subsequent factorization.

To demonstrate, let's consider a simplified CSL comprised of a single three-component reaction with 10,000 distinct synthons for each R-group, i.e., $|S|=30,000$. This yields a total of one trillion products in $\mathcal{D}$. Exhaustive screening with $\hat{f}_\theta$ would therefore require one trillion neural network evaluations. APEX, on the other hand, produces all intermediate representations for the library with just 30,000 neural network evaluations. The associative embeddings \eqref{assoc_embeds} can be cached as a $|S| \times d$ matrix for later re-use. Supposing $d=1024$, this would require about 120 MB of memory. In contrast, to cache the latent representations for all of the one trillion products in $\mathcal{D}$ would require about 4 PB of memory. With the associative embeddings in our possession, we can calculate their dot products with the task weight ${\bf w}_i$, which is just $2|S| \times d - |S|=61.41$ million floating point operations. Once these terms have been computed, each $\hat{\hat f}_i({\bf x})$ can be calculated with just a few floating point operations (three in this case: the summation of the three synthon associative contributions and the bias term $b_i$). Hence, we can approximate the surrogate predictions for all compounds in $\mathcal{D}$ with just three trillion floating point operations (i.e., 3 TFLOP). Noting that the NVIDIA Tesla T4 GPU is able to perform 8.1 TFLOPS, the APEX factorization \eqref{apex} theoretically permits evaluation of all one trillion compounds in the CSL in just a few seconds.

To construct the top-$k$ set $\hat{\hat{X}}_k^*$ for the retrieval problem \eqref{apex_topk}, we can stream the pre-computed synthon associative contributions for each task and add them with the bias to form the APEX prediction \eqref{apex}. We can compute the constraint violation under the APEX predictions,
\begin{align}
    \hat{\hat{c}}({\bf x}) &= -\sum_{i=1}^m\max(0, \ell_i - \hat{\hat{f}}_i({\bf x})) \nonumber \\
    &\qquad - \sum_{i=1}^m\max(0, \hat{\hat{f}}_i({\bf x}) - u_i),
\end{align}
which is zero if all of the predicted constraints are satisfied and negative if there is any violation. Hence, for each compound in $\mathcal{D}$, we form a two-dimensional vector $(\hat{\hat{c}}({\bf x}), \hat{\hat{f}}_0({\bf x}))$ that is used to enter compounds into a priority queue of size $k$ that organizes them in lexicographical order. Once we have exhausted through all compounds in $\mathcal{D}$, we can remove any compound from the top-$k$ set where $\hat{\hat{c}}({\bf x})<0$. The result is the solution $\hat{\hat X}_k^*$ to \eqref{apex_topk}.

\subsection{Top-\emph{k} Retrieval}

The exposition in the previous subsection on runtime only considers evaluation of APEX predictions over the entire CSL and ignores overhead introduced by maintenance of the top-$k$ set. APEX implements custom top-$k$ algorithms for the CPU and GPU within a PyTorch CUDA C++ extension module. The CPU PyTorch operator calculates each molecule's APEX objective $\hat{\hat{f}}_0({\bf x})$ and constraint violation $\hat{\hat{c}}({\bf x})$ on the fly and streams them directly into a priority queue. However, APEX is uniquely suited for the GPU, as it requires only a small initial data transfer from CPU to GPU, and all intermediate calculations can be performed entirely on-GPU. To leverage the high compute capability and memory bandwidth of the GPU, the CUDA PyTorch operator employs a chain-of-batches strategy with the GPU-compatible AIR top-$k$ algorithm \citep{zhang2023topk}. Additional details are provided in Appendix \ref{sec:gputopk}.

\section{Evaluation}

\begin{figure*}[!ht]
  \centering
  \includegraphics[width=\textwidth]{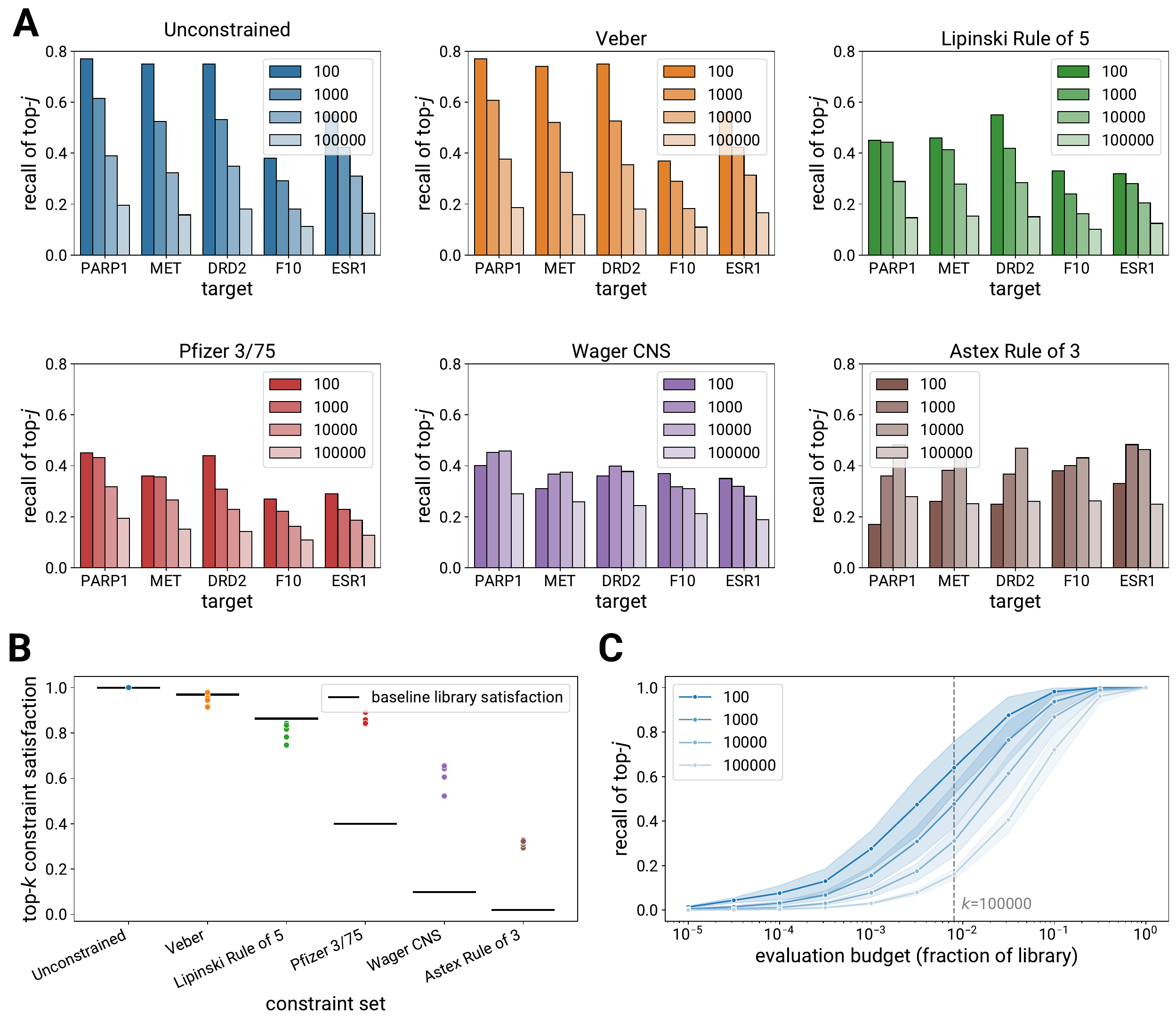}
  \caption{(A) Percent of compounds in the ground truth top-$j$ set retrieved by the APEX top $k=\text{100,000}$ set from the 12M compound CSL. A random baseline will achieve a recall below 0.01. (B) Constraint satisfaction rates for the APEX retrievals. Black line denotes the base fraction of satisfying compounds in the library for each set of constraints. (C) Recall of different top-$j$ sets without constraints as a function of increasing evaluation budget. Recall is averaged over all five targets, with error bars showing the standard deviation. Per-target recall curves are shown in Figure \ref{fig:recall_at_k} of the Appendix. The dashed line corresponds to $k=\text{100,000}$, the budget used for (A) and (B).}
  \label{fig:recall}
\end{figure*}

To demonstrate APEX's capabilities on a variety of pertinent virtual screening queries, we evaluate its ability to accurately retrieve the top-$k$ compounds in a large, representative CSL by docking score across the five selected targets (PARP1, MET, DRD2, F10, and ESR1) and against a number of relevant constraint sets used in drug discovery (described in Appendix \ref{constraints}).

In all reported experiments, the surrogate is trained on the 1M compound CSL described in Section \ref{sec:data} (this is the only step in which labels are provided to the model) and the factorizer is trained on either the 12M or 10B compound CSL (in the absence of labels) to reconstruct embeddings produced by the trained surrogate model. We use an embedding dimension of $d=64$. No extensive hyperparameter tuning was performed; we opted for a lightweight model for purposes of demonstrating APEX (but note that runtime for APEX search is not a function of $d$ once pre-calculations have been performed).



\subsection{Top-$k$ Retrieval}

For a library $\mathcal{D}$, objective $f_0$, constraints $\{(f_i,\ell_i, u_i)\}_{i=1}^m$, and evaluation budget $k$, we are ultimately interested in a screening algorithm's ability to accurately retrieve the ground truth top-$j$ set $X^*_j$ in \eqref{orig_topk}, where $j\le k$. For example, we might have the budget to evaluate $k=\text{100,000}$ compounds but wish to quantify what percent of the top-$j=\text{100}$ were correctly retrieved. For APEX, this quantity can be expressed simply as
\begin{align}
    \text{Recall-$j$-at-$k$} &= \frac{|X_{j}^*\cap \hat{X}_{k}^*|}{|X_{j}^*|}.
\end{align}
Of course, this requires knowing the ground truth top-$j$ set $X_j^*$ for a given search query. We use the 12M enumerated and exhaustively scored CSL to perform such an evaluation.

Results are shown in Figure \ref{fig:recall}A. With a budget of $k=\text{100,000}$ retrievals (representing 0.803\% of compounds in the CSL), the ground truth top-$j$ compounds are recovered at rates far exceeding selection with a random baseline across all targets and for all sets of constraints. 
In addition to the search without constraints, the recall is highest for the Veber set of constraints, which are the least stringent and are satisfied by most compounds in the library (Figure \ref{fig:recall}B).
The Astex Rule-of-3 constraints are designed for fragment-based drug discovery but we include them here as an example of a more stringent constraint set. While the rate of constraint satisfaction is lowest for this set, it is still much higher than the baseline rate of constraint satisfaction in the library.



\subsection{Comparison with Thompson Sampling}
We compare APEX against the Thompson sampling (TS) algorithm of \citet{klarich2024thompson}, which was specifically designed for screening ultra-large CSLs. Thompson sampling \citep{thompson1933likelihood} is a classic Bayesian decision-making strategy in which a belief distribution over expected rewards is maintained for each action, an action is selected by sampling from these beliefs, and the distribution is updated upon observing the outcome. In the context of CSL screening, \citet{klarich2024thompson} instantiate this principle in synthon space: for each reaction, independent belief distributions are maintained over the expected reward contribution of each synthon, synthon combinations are sampled according to these beliefs and evaluated with the oracle, and the distributions are updated accordingly.

Crucially, this TS implementation operates separately on each reaction, i.e., the multi-armed bandit is defined at the level of a single reaction's synthon pool, and results are aggregated across reactions post hoc, which limits its ability to jointly optimize over the full library. As TS is run on each reaction separately, we limit the comparison to the top five largest reactions in our 12M CSL (in total comprising over 4 million products). The total number of evaluations for TS is $|S| \times w + i$, where $S$ is the set of synthons for that particular reaction, $w$ is the number of warmup steps, and $i$ is the number of TS iterations and output molecules.

We run TS for 100, 1{,}000, or 10{,}000 iterations, with 3 warmup steps for two-component reactions and 10 for three-component reactions (as suggested in \citet{klarich2024thompson}). As this TS implementation does not directly support constraints on molecular properties, we perform the comparison in the unconstrained case for both APEX and TS, only minimizing docking score as the objective. Full results are shown in Figure~\ref{fig:ts} in the Appendix. While results vary across targets and reactions, APEX consistently outperforms or matches TS at recalling the top-$j$ compounds, showing particular strength at lower evaluation budgets.

\subsection{Docking Score Enrichment on Ultra-Large Libraries}

Figure \ref{fig:scores} plots the empirical CDF of docking scores across the five targets for the APEX top-$k$ set in both the 10B and 12M compound CSLs against the background distribution of scores from the 12M compound library. This result demonstrates clear enrichment in the APEX top-$k$ sets relative to the background set, and further highlights the value of screening larger CSLs to identify higher scoring compounds enabled by APEX's accelerated runtime and ability to scale to ultra-large combinatorial libraries.

\subsubsection{Zero-Shot Application to the Enamine REAL Library}

In addition to the BRICS CSLs, we also apply APEX to the commercial Enamine REAL library (9-2024 version). This library contains more than 70B compounds and serves as a test of APEX's generality, both in scaling to even larger library sizes and as an application of a pretrained surrogate and factorizer in a zero-shot manner.

Figure \ref{fig:scores} presents the docking score distributions from this library alongside a background score distribution generated from 100,000 random compounds.
Despite the surrogate and factorizer being trained on a different, much smaller library, APEX is able to enrich docking scores with respect to the background distribution of the Enamine library and, in most cases, with respect to the top-$k$ of the 10B BRICS library. The lowest enrichment is from MET kinase, which also corresponds to the largest drop-off in R-squared in this zero-shot application of the factorizer (Figure \ref{fig:r2} in the Appendix). While these results demonstrate APEX's capabilities in a zero-shot context on ultra-large vendor CSLs today, even greater docking score enrichment is likely achievable through fine-tuning the surrogate (and subsequently the factorizer) using labeled data from the target CSL.

\subsection{Execution Speed of APEX on Ultra-Large CSLs}

Table \ref{tab:runtime} reports runtimes of APEX top-$k$ search on the BRICS and Enamine libraries for different choices of $k$, evaluated on a single NVIDIA Tesla T4 GPU. The reported runtimes represent end-to-end execution, i.e., from problem specification to an output dataframe with APEX top-$k$ SMILES and their associated APEX-predicted objective and constraint values. In screening the 10B and 70B compound libraries, we observe an order of magnitude speedup in runtime when using the GPU top-$k$ implementation as opposed to CPU. Further, as constraints are included, the gap widens significantly, with the CPU implementation's runtime increasing approximately linearly in the number of constraints added. Using the GPU top-$k$ implementation, APEX is able to retrieve the approximate top $k=\text{1,000,000}$ compounds from a 10B compound library in less than thirty seconds under standard drug likeness constraints, making it a highly performant and scalable search protocol for ultra-large CSLs. 


\begin{table}[h]
\centering
\small
\begin{tabular}{rccc}
\toprule
\multicolumn{4}{c}{\textbf{Unconstrained}} \\
\midrule
$k$ & BRICS 12M & BRICS 10B & REAL 70B \\
\midrule
10,000    & 0.3 (0.4)    & 10.9 (130.7) & 168.4 (838.5) \\
100,000   & 1.2 (1.2)    & 11.6 (131.4) & 169.3 (847.2) \\
1,000,000 & 10.9 (12.7)  & 21.2 (147.6) & 184.0 (858.9) \\
\midrule
\midrule
\multicolumn{4}{c}{\textbf{Lipinski Rule of 5}} \\
\midrule
$k$ & BRICS 12M & BRICS 10B & REAL 70B \\
\midrule
10,000    & 0.3 (0.7)    & 13.9 (437.7) & 186.1 (3163.2) \\
100,000   & 0.9 (1.7)    & 14.6 (443.7) & 187.5 (3184.6) \\
1,000,000 & 10.8 (12.9)  & 24.3 (462.2) & 202.4 (3142.5) \\
\bottomrule
\end{tabular}
\caption{Runtime of APEX top-$k$ search across constraints and library sizes in seconds. Times are averaged over five runs (one with each target's docking score as an objective), with GPU runtime reported first and CPU runtime reported in parentheses.}
\label{tab:runtime}
\end{table}

\section{Conclusion}

In this paper, we proposed the APEX search protocol for the virtual screening of combinatorial synthesis libraries, enabling the rapid execution of declarative queries that scales to ultra-large libraries (in excess of 10 billion compounds). While traditional virtual screening algorithms are limited by design to evaluate only a small fraction of the eligible search space, APEX enables a fast, exhaustive evaluation over the entire search space by taking advantage of the structure of CSLs. This allows researchers to rapidly identify high-scoring compounds virtually that satisfy design constraints. We demonstrated APEX's capabilities on a benchmark CSL of over 10 million compounds, all annotated with ground truth docking scores and physicochemical properties. Our results show that APEX consistently achieves high recall rates for the ground truth top-$k$ compounds at low $k$ and effectively satisfies diverse constraint sets, far exceeding random baselines.

APEX is a significant step towards making exhaustive virtual screening a routine computational task. Its ability to efficiently screen entire CSLs ensures that valuable, high-scoring compounds are not overlooked. Moreover, due to its rapid execution speed---virtually screening a CSL in excess of 10 billion compounds in less than 30 seconds with a single Tesla T4 GPU---APEX enables rapid hypothesis testing and interactive exploration of chemical space. 

\section*{Acknowledgments}
The authors thank David Graff and Henry van den Bedem for early conceptual input, Yurii Moroz for feedback on library construction, and Kate Stafford for discussions on the manuscript, experiments, and applications.

\section*{Impact Statement}

This paper introduces the APEX search protocol to enable the rapid, exhaustive evaluation of ultra-large combinatorial synthesis libraries (CSLs) containing tens of billions of compounds. This technology lowers the computational barriers to exploring such massive, make-on-demand chemical spaces, potentially accelerating the identification of novel therapeutic agents.


\bibliography{references}

@article{garcia2022dockstring,
  title={{DOCKSTRING}: Easy molecular docking yields better benchmarks for ligand design},
  author={Garc{\'\i}a-Orteg{\'o}n, Miguel and Simm, Gregor NC and Tripp, Austin J and Hern{\'a}ndez-Lobato, Jos{\'e} Miguel and Bender, Andreas and Bacallado, Sergio},
  journal={Journal of Chemical Information and Modeling},
  volume={62},
  number={15},
  pages={3486--3502},
  year={2022},
  publisher={ACS Publications}
}

@article{pedawi2022efficient,
  title={An efficient graph generative model for navigating ultra-large combinatorial synthesis libraries},
  author={Pedawi, Aryan and Gniewek, Pawel and Chang, Chaoyi and Anderson, Brandon and van den Bedem, Henry},
  journal={Advances in Neural Information Processing Systems},
  volume={35},
  pages={8731--8745},
  year={2022}
}

@article{de2024ngt,
  title={{NGT}: Generative {AI} with synthesizability guarantees discovers {MC2R} inhibitors from a tera-scale virtual screen},
  author={de Oliveira, Saulo HP and Pedawi, Aryan and Kenyon, Victor and van den Bedem, Henry},
  journal={Journal of Medicinal Chemistry},
  volume={67},
  number={21},
  pages={19417--19427},
  year={2024},
  publisher={ACS Publications}
}

@inproceedings{pedawi2023through,
  title={Through the looking glass: Navigating in latent space to optimize over combinatorial synthesis libraries},
  author={Pedawi, Aryan and De Oliveira, Saulo and van den Bedem, Henry},
  booktitle={NeurIPS 2023 Generative AI and Biology (GenBio) Workshop},
  url={https://openreview.net/forum?id=ALsSka1db3},
  year={2023}
}

@inproceedings{zhang2023topk,
author = {Zhang, Jingrong and Naruse, Akira and Li, Xipeng and Wang, Yong},
title = {Parallel Top-K Algorithms on {GPU}: A Comprehensive Study and New Methods},
year = {2023},
isbn = {9798400701092},
publisher = {Association for Computing Machinery},
address = {New York, NY, USA},
url = {https://doi.org/10.1145/3581784.3607062},
doi = {10.1145/3581784.3607062},
booktitle = {Proceedings of the International Conference for High Performance Computing, Networking, Storage and Analysis},
articleno = {76},
numpages = {13},
location = {Denver, CO, USA},
series = {SC '23}
}

@article{sadybekov2022synthon,
  title={Synthon-based ligand discovery in virtual libraries of over 11 billion compounds},
  author={Sadybekov, Arman A and Sadybekov, Anastasiia V and Liu, Yongfeng and Iliopoulos-Tsoutsouvas, Christos and Huang, Xi-Ping and Pickett, Julie and Houser, Blake and Patel, Nilkanth and Tran, Ngan K and Tong, Fei and others},
  journal={Nature},
  volume={601},
  number={7893},
  pages={452--459},
  year={2022},
  publisher={Nature Publishing Group UK London}
}

@article{klarich2024thompson,
  title={Thompson Sampling: An Efficient Method for Searching Ultra-large Synthesis on Demand Databases},
  author={Klarich, Kathryn and Goldman, Brian and Kramer, Trevor and Riley, Patrick and Walters, W Patrick},
  journal={Journal of Chemical Information and Modeling},
  year={2024},
  publisher={ACS Publications}
}

@article{degen2008art,
  title={On the art of compiling and using ``drug-like'' chemical fragment spaces},
  author={Degen, Jorg and Wegscheid-Gerlach, Christof and Zaliani, Andrea and Rarey, Matthias},
  journal={ChemMedChem},
  volume={3},
  number={10},
  pages={1503},
  year={2008}
}

@article{tingle2023zinc,
  title={{ZINC}-22: A free multi-billion-scale database of tangible compounds for ligand discovery},
  author={Tingle, Benjamin I and Tang, Khanh G and Castanon, Mar and Gutierrez, John J and Khurelbaatar, Munkhzul and Dandarchuluun, Chinzorig and Moroz, Yurii S and Irwin, John J},
  journal={Journal of Chemical Information and Modeling},
  volume={63},
  number={4},
  pages={1166--1176},
  year={2023},
  publisher={ACS Publications}
}

@misc{landrum2006rdkit,
  title={RDKit: Open-source cheminformatics},
  author={Landrum, Greg and others},
  year={2006},
  publisher={Zenodo}
}

@article{trott2010autodock,
  title={{A}uto{D}ock {V}ina: Improving the speed and accuracy of docking with a new scoring function, efficient optimization, and multithreading},
  author={Trott, Oleg and Olson, Arthur J},
  journal={Journal of Computational Chemistry},
  volume={31},
  number={2},
  pages={455--461},
  year={2010},
  publisher={Wiley Online Library}
}

@article{hughes2008physiochemical,
  title={Physiochemical drug properties associated with in vivo toxicological outcomes},
  author={Hughes, Jason D and Blagg, Julian and Price, David A and Bailey, Simon and DeCrescenzo, Gary A and Devraj, Rajesh V and Ellsworth, Edmund and Fobian, Yvette M and Gibbs, Michael E and Gilles, Richard W and others},
  journal={Bioorganic \& Medicinal Chemistry Letters},
  volume={18},
  number={17},
  pages={4872--4875},
  year={2008},
  publisher={Elsevier}
}

@article{congreve2003rule,
  title={A ``rule of three'' for fragment-based lead discovery?},
  author={Congreve, Miles and Carr, Robin and Murray, Chris and Jhoti, Harren},
  journal={Drug Discovery Today},
  volume={8},
  number={19},
  pages={876--877},
  year={2003}
}

@article{veber2002molecular,
  title={Molecular properties that influence the oral bioavailability of drug candidates},
  author={Veber, Daniel F and Johnson, Stephen R and Cheng, Hung-Yuan and Smith, Brian R and Ward, Keith W and Kopple, Kenneth D},
  journal={Journal of Medicinal Chemistry},
  volume={45},
  number={12},
  pages={2615--2623},
  year={2002},
  publisher={ACS Publications}
}

@article{lipinski1997experimental,
  title={Experimental and computational approaches to estimate solubility and permeability in drug discovery and development settings},
  author={Lipinski, Christopher A and Lombardo, Franco and Dominy, Beryl W and Feeney, Paul J},
  journal={Advanced Drug Delivery Reviews},
  volume={23},
  number={1-3},
  pages={3--25},
  year={1997},
  publisher={Elsevier}
}

@inproceedings{morrison2020cuina,
  title={{CU}ina: An Efficient {GPU} Implementation of {A}uto{D}ock {V}ina},
  author={Morrison, Adrian and Friedland, Greg and Wallach, Izhar},
  booktitle={American Chemical Society Fall 2020 Virtual Meeting \& Expo},
  month={August},
  year={2020},
}

@manual{cccl,
  title = {{CCCL}: {CUDA} {C++} {C}ore {L}ibraries},
  author = {{CCCL Development Team}},
  year = {2023},
  url = {https://github.com/NVIDIA/cccl},
}

@article{griliches1974errors,
  title={Errors in variables and other unobservables},
  author={Griliches, Zvi},
  journal={Econometrica: Journal of the Econometric Society},
  pages={971--998},
  year={1974},
  publisher={JSTOR}
}

@article{wager2010moving,
  title={Moving beyond rules: the development of a central nervous system multiparameter optimization (CNS MPO) approach to enable alignment of druglike properties},
  author={Wager, Travis T and Hou, Xinjun and Verhoest, Patrick R and Villalobos, Anabella},
  journal={ACS chemical neuroscience},
  volume={1},
  number={6},
  pages={435--449},
  year={2010},
  publisher={ACS Publications}
}

@article{graff2021accelerating,
  title={Accelerating high-throughput virtual screening through molecular pool-based active learning},
  author={Graff, David E and Shakhnovich, Eugene I and Coley, Connor W},
  journal={Chemical science},
  volume={12},
  number={22},
  pages={7866--7881},
  year={2021},
  publisher={Royal Society of Chemistry}
}

@article{mehta2021memes,
  title={Memes: Machine learning framework for enhanced molecular screening},
  author={Mehta, Sarvesh and Laghuvarapu, Siddhartha and Pathak, Yashaswi and Sethi, Aaftaab and Alvala, Mallika and Priyakumar, U Deva},
  journal={Chemical science},
  volume={12},
  number={35},
  pages={11710--11721},
  year={2021},
  publisher={Royal Society of Chemistry}
}

@article{luo2024projecting,
  title={Projecting molecules into synthesizable chemical spaces},
  author={Luo, Shitong and Gao, Wenhao and Wu, Zuofan and Peng, Jian and Coley, Connor W and Ma, Jianzhu},
  journal={arXiv preprint arXiv:2406.04628},
  year={2024}
}

@article{gao2025generative,
  title={Generative AI for navigating synthesizable chemical space},
  author={Gao, Wenhao and Luo, Shitong and Coley, Connor W},
  journal={Proceedings of the National Academy of Sciences},
  volume={122},
  number={41},
  pages={e2415665122},
  year={2025},
  publisher={National Academy of Sciences}
}

@article{gentile2020deep,
  title={Deep docking: a deep learning platform for augmentation of structure based drug discovery},
  author={Gentile, Francesco and Agrawal, Vibudh and Hsing, Michael and Ton, Anh-Tien and Ban, Fuqiang and Norinder, Ulf and Gleave, Martin E and Cherkasov, Artem},
  journal={ACS central science},
  volume={6},
  number={6},
  pages={939--949},
  year={2020},
  publisher={ACS Publications}
}

@article{graff2022self,
  title={Self-focusing virtual screening with active design space pruning},
  author={Graff, David E and Aldeghi, Matteo and Morrone, Joseph A and Jordan, Kirk E and Pyzer-Knapp, Edward O and Coley, Connor W},
  journal={Journal of Chemical Information and Modeling},
  volume={62},
  number={16},
  pages={3854--3862},
  year={2022},
  publisher={ACS Publications}
}

@inproceedings{cretu2024synflownet,
  title={Synflownet: Towards molecule design with guaranteed synthesis pathways},
  author={Cretu, Miruna and Harris, Charles and Roy, Julien and Bengio, Emmanuel and Li{\`o}, Pietro},
  booktitle={ICLR 2024 Workshop on Generative and Experimental Perspectives for Biomolecular Design},
  year={2024}
}

@inproceedings{gao2023drugclip,
  title={{DrugCLIP}: Contrastive protein-molecule representation learning for virtual screening},
  author={Gao, Bowen and Qiang, Bo and Tan, Haichuan and Jia, Yinjun and Ren, Minsi and Lu, Minsi and Liu, Jingjing and Ma, Wei-Ying and Lan, Yanyan},
  booktitle={Advances in Neural Information Processing Systems},
  volume={36},
  year={2023}
}

@article{gao2026drugclip,
  title={Deep contrastive learning enables genome-wide virtual screening},
  author={Gao, Bowen and others},
  journal={Science},
  year={2026},
  doi={10.1126/science.ads9530}
}

@article{singh2023conplex,
  title={Contrastive learning in protein language space predicts interactions between drugs and protein targets},
  author={Singh, Rohit and Sledzieski, Samuel and Bryson, Bryan and Cowen, Lenore and Berger, Bonnie},
  journal={Proceedings of the National Academy of Sciences},
  volume={120},
  number={24},
  pages={e2220778120},
  year={2023},
  publisher={National Academy of Sciences}
}

@article{feng2025ligunity,
  title={Hierarchical affinity landscape navigation through learning a shared pocket-ligand space},
  author={Feng, Bin and Liu, Zijing and Li, Hao and Yang, Mingjun and Zou, Junjie and Cao, He and Li, Yu and Zhang, Lei and Wang, Sheng},
  journal={Patterns},
  volume={6},
  pages={101371},
  year={2025},
  publisher={Elsevier},
  doi={10.1016/j.patter.2025.101371}
}

@inproceedings{rendle2010fm,
  title={Factorization machines},
  author={Rendle, Steffen},
  booktitle={2010 IEEE International Conference on Data Mining (ICDM)},
  pages={995--1000},
  year={2010},
  organization={IEEE},
  doi={10.1109/ICDM.2010.127}
}

@article{koren2009matrix,
  title={Matrix factorization techniques for recommender systems},
  author={Koren, Yehuda and Bell, Robert and Volinsky, Chris},
  journal={IEEE Computer},
  volume={42},
  number={8},
  pages={30--37},
  year={2009},
  publisher={IEEE}
}

@article{thompson1933likelihood,
  title={On the likelihood that one unknown probability exceeds another 
         in view of the evidence of two samples},
  author={Thompson, William R},
  journal={Biometrika},
  volume={25},
  number={3-4},
  pages={285--294},
  year={1933},
  publisher={Oxford University Press}
}

@article{bickerton2012quantifying,
  title={Quantifying the chemical beauty of drugs},
  author={Bickerton, G Richard and Paolini, Gaia V and Besnard, J{\'e}r{\'e}my and Muresan, Sorel and Hopkins, Andrew L},
  journal={Nature Chemistry},
  volume={4},
  number={2},
  pages={90--98},
  year={2012},
  publisher={Nature Publishing Group}
}

@article{kosugi2021quantitative,
  title={Quantitative estimate of protein-protein interaction inhibitor accessibility of compounds: {QEPPI}},
  author={Kosugi, Tatsuya and Ohue, Masahito},
  journal={Journal of Chemical Information and Modeling},
  volume={61},
  number={2},
  pages={718--730},
  year={2021},
  publisher={ACS Publications}
}

@article{ertl2009estimation,
  title={Estimation of synthetic accessibility score of drug-like molecules based on molecular complexity and fragment contributions},
  author={Ertl, Peter and Schuffenhauer, Ansgar},
  journal={Journal of Cheminformatics},
  volume={1},
  number={1},
  pages={8},
  year={2009},
  publisher={Springer}
}

@article{krzyzanowski2023sps,
  title={Spacial score---a comprehensive topological indicator for small-molecule complexity},
  author={Krzyzanowski, Adrian and Pahl, Axel and Grigalunas, Michael and Waldmann, Herbert},
  journal={Journal of Medicinal Chemistry},
  volume={66},
  number={18},
  pages={12739--12750},
  year={2023},
  publisher={ACS Publications},
  doi={10.1021/acs.jmedchem.3c00689}
}
\bibliographystyle{icml2026}

\newpage
\appendix
\onecolumn
\section{Appendix}

\subsection{Combinatorial Synthesis Libraries}\label{app:csl_structure}

A combinatorial synthesis library (CSL), $\mathcal{D} \equiv (\mathcal{T}, \mathcal{R}, \mathcal{S}, \psi, \varphi, \phi)$, is organized hierarchically across three levels: synthons $\mathcal{S}$ at the bottom, R-groups $\mathcal{R}$ in the middle, and reactions $\mathcal{T}$ at the top.

\textbf{Synthons.}
A synthon $s \in \mathcal{S}$ is a molecular building block, i.e., a molecular fragment with annotated attachment points, represented as a molecular graph or SMILES string $x_s \in \mathcal{X}^*$, where $\mathcal{X}^* \supset \mathcal{X}$ extends the space of complete molecules to include attachment point tokens.

\textbf{R-groups.}
An R-group $r \in \mathcal{R}$ is a placeholder component in a multi-component reaction, spanned by a set of eligible synthons. The function $\psi_{\mathcal{R} \to \mathcal{S}} : \mathcal{R} \to \mathcal{P}(\mathcal{S})$ returns the set of synthons $\psi_{\mathcal{R} \to \mathcal{S}}(r) \subset \mathcal{S}$ that are valid assignments for a given R-group $r$, where $\mathcal{P}(\cdot)$ denotes the power set.

\textbf{Reactions.}
A reaction $t \in \mathcal{T}$ is a multi-component synthesis rule that combines one synthon per R-group to produce a single product molecule.
The function $\psi_{\mathcal{T} \to \mathcal{R}} : \mathcal{T} \to \mathcal{P}(\mathcal{R})$ returns the set of R-groups $\psi_{\mathcal{T} \to \mathcal{R}}(t) \subset \mathcal{R}$ associated with reaction $t$.

Each product in $\mathcal{D}$ is uniquely identified by a multi-index $\chi = (t, \{(r, s) : s \in \psi_{\mathcal{R} \to \mathcal{S}}(r),\ \forall r \in \psi_{\mathcal{T} \to \mathcal{R}}(t)\})$ describing its reaction and R-group assignment, and the synthesis rule $\phi$ maps this multi-index to a product molecule $x := \phi(\chi)$.

Due to the combinatorial design of CSLs, the number of products grows multiplicatively with the number of synthons per R-group. For example, a library containing a single three-component reaction with 10,000 synthons per R-group enumerates a chemical space of one trillion products from only 30,000 synthons ($10,000^3 = 10^{12}$). Commercially available CSLs like Enamine REAL achieve tens of billions of synthetically accessible products from a few hundred thousand synthons distributed across hundreds of reactions.

\textbf{Encoder hierarchy.}
The APEX factorizer mirrors this three-level hierarchy with a corresponding stack of neural network encoders (see Section~3.2 for full details). At the bottom, the \texttt{SynthonEncoder} processes the molecular graph of each synthon $s$ to produce a synthon-level embedding:
\begin{equation}
    h^{\mathcal{S}}_s = \texttt{SynthonEncoder}(x_s).
\end{equation}
A deep set network, the \texttt{RgroupEncoder}, then aggregates the synthon embeddings belonging to each R-group $r$:
\begin{equation}
    h^{\mathcal{R}}_r = \texttt{RgroupEncoder}(\{h^{\mathcal{S}}_s : \forall s \in \psi_{\mathcal{R} \to \mathcal{S}}(r)\}).
\end{equation}
Finally, another deep set network, the \texttt{ReactionEncoder}, aggregates the R-group embeddings belonging to each reaction $t$:
\begin{equation}
    h^{\mathcal{T}}_t = \texttt{ReactionEncoder}(\{h^{\mathcal{R}}_r : \forall r \in \psi_{\mathcal{T} \to \mathcal{R}}(t)\}).
\end{equation}
These representations form the basis for the \texttt{SynthonValueEncoder} and \texttt{RgroupKeyEncoder} that produce the synthon associative contributions $v_{i,r,s}$ used in the APEX factorization (\eqref{apex}). Once computed, these contributions are cached for all synthons in the library, so that scoring any compound at search time requires only a small, fixed number of additions regardless of library size. This encoding procedure scales as $\mathcal{O}(|\mathcal{S}| + |\mathcal{R}| + |\mathcal{T}|)$, which is logarithmic relative to the $\mathcal{O}(|\mathcal{D}|)$ cost of encoding each product individually, and is performed once per library.

We refer the reader to \citet{pedawi2022efficient} for additional background.

\subsection{The Virtual Library}

\begin{figure}[ht]
  \begin{center}
  \includegraphics[width=\textwidth]{imgs/csl.pdf}
  \end{center}
  \caption{Beyond commercially available make-on-demand CSLs, it is relatively straightforward to design an ultra-large CSL for virtual screening using publicly available libraries of enumerated compounds like ZINC22 and cheminformatics tools like RDKit. These designs are incredibly valuable for virtual screening due to their ability to densely cover large swaths of relevant chemical space.}
  \label{fig:csl}
\end{figure}

\begin{figure}[ht]
  \centering
  \includegraphics[width=\textwidth]{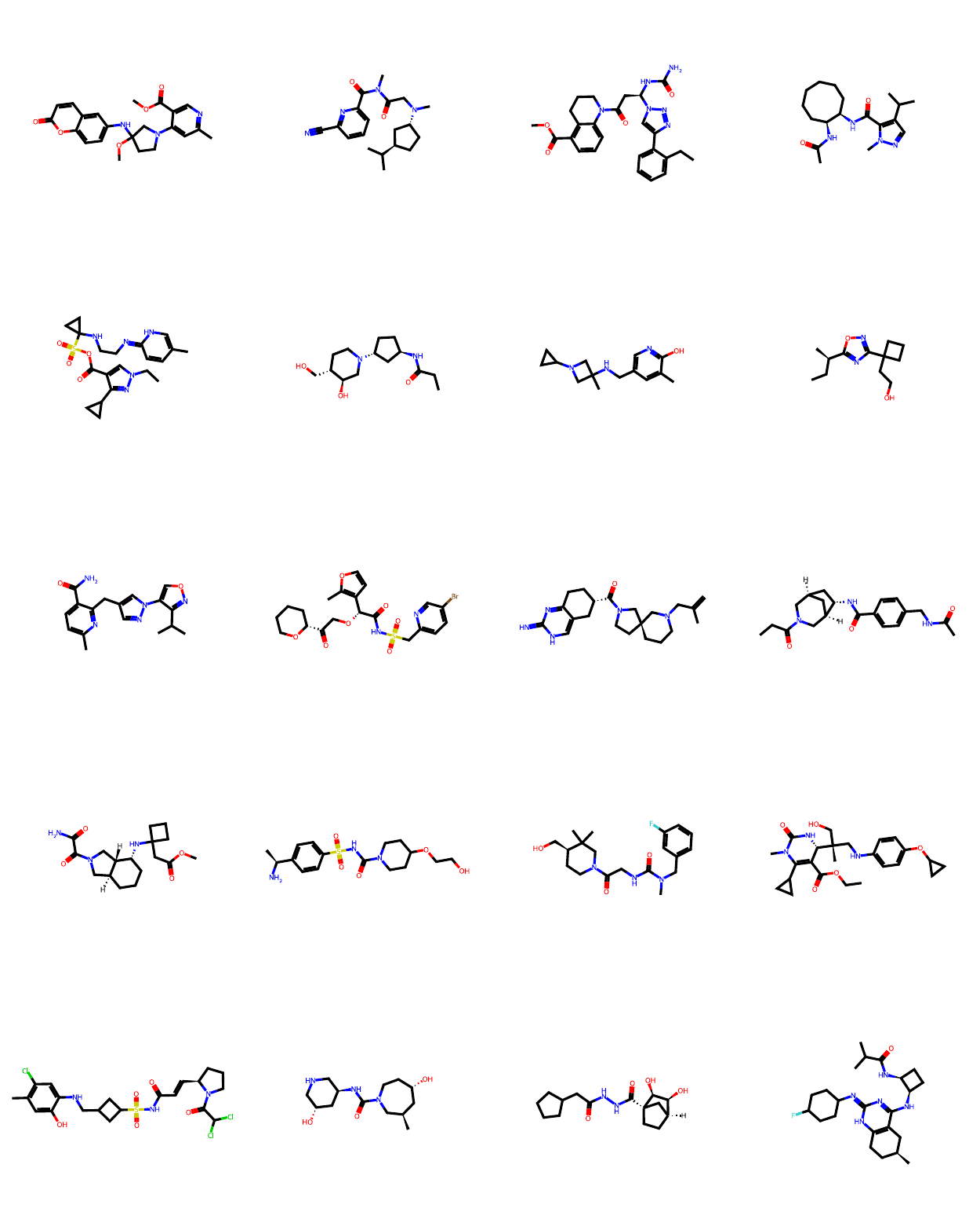}
  \caption{Example molecules from the 10B compound CSL.}
  \label{fig:example_molecules}
\end{figure}

\begin{figure}[ht]
  \centering
  \includegraphics[width=\textwidth]{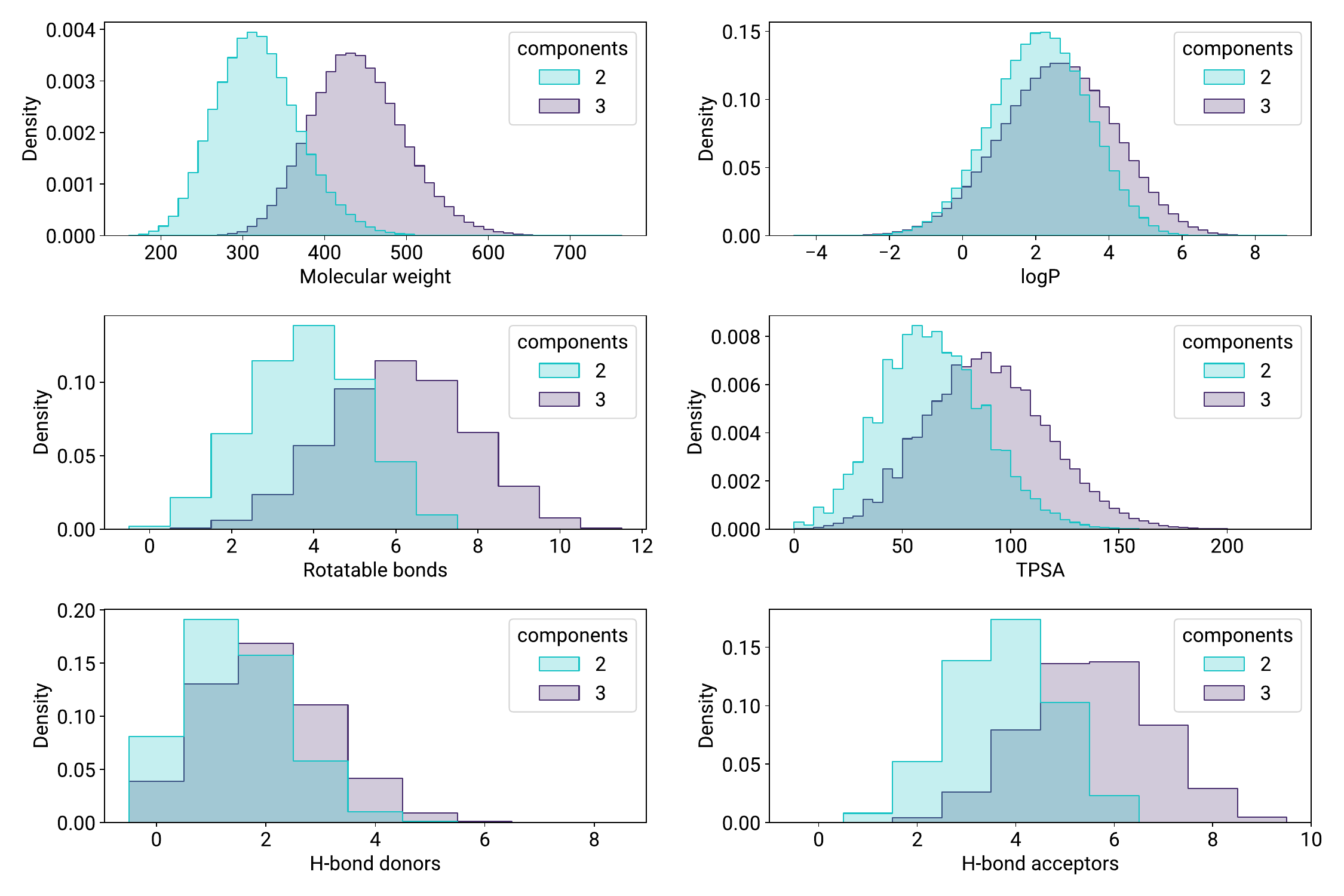}
  \caption{Distribution of select molecular properties in the 12M compound CSL.}
  \label{fig:property_distribution}
\end{figure}

Figure \ref{fig:example_molecules} displays twenty randomly selected molecules from the 10B compound CSL constructed as part of this study. In Figure \ref{fig:property_distribution}, the distribution of molecular properties for the fully enumerated 12M compound CSL are shown for both two- and three-component reactions. We note that compounds originating from three-component reactions tend to be larger than those from two-component reactions.

\subsection{Description of Computational Endpoints}

Table \ref{tab:endpoints} provides a full list of the molecular property and docking score endpoints considered in this study.

Docking scores against the five protein targets (PARP1, MET, DRD2, F10, and ESR1) were computed using CUina \citep{morrison2020cuina}, an efficient GPU implementation of AutoDock Vina \citep{trott2010autodock}, with receptor structures and binding site definitions taken from the DOCKSTRING dataset \citep{garcia2022dockstring}. The physicochemical properties were computed using RDKit \citep{landrum2006rdkit} or RDKit-based calculators.

All endpoints were computed on the fully enumerated 1M and 12M compound BRICS-based CSLs and are used throughout this study both as optimization objectives and as constraints in virtual screening queries.

\subsection{Regression Performance by Endpoint}

\begin{figure}[ht]
  \centering
  \includegraphics[width=\textwidth]{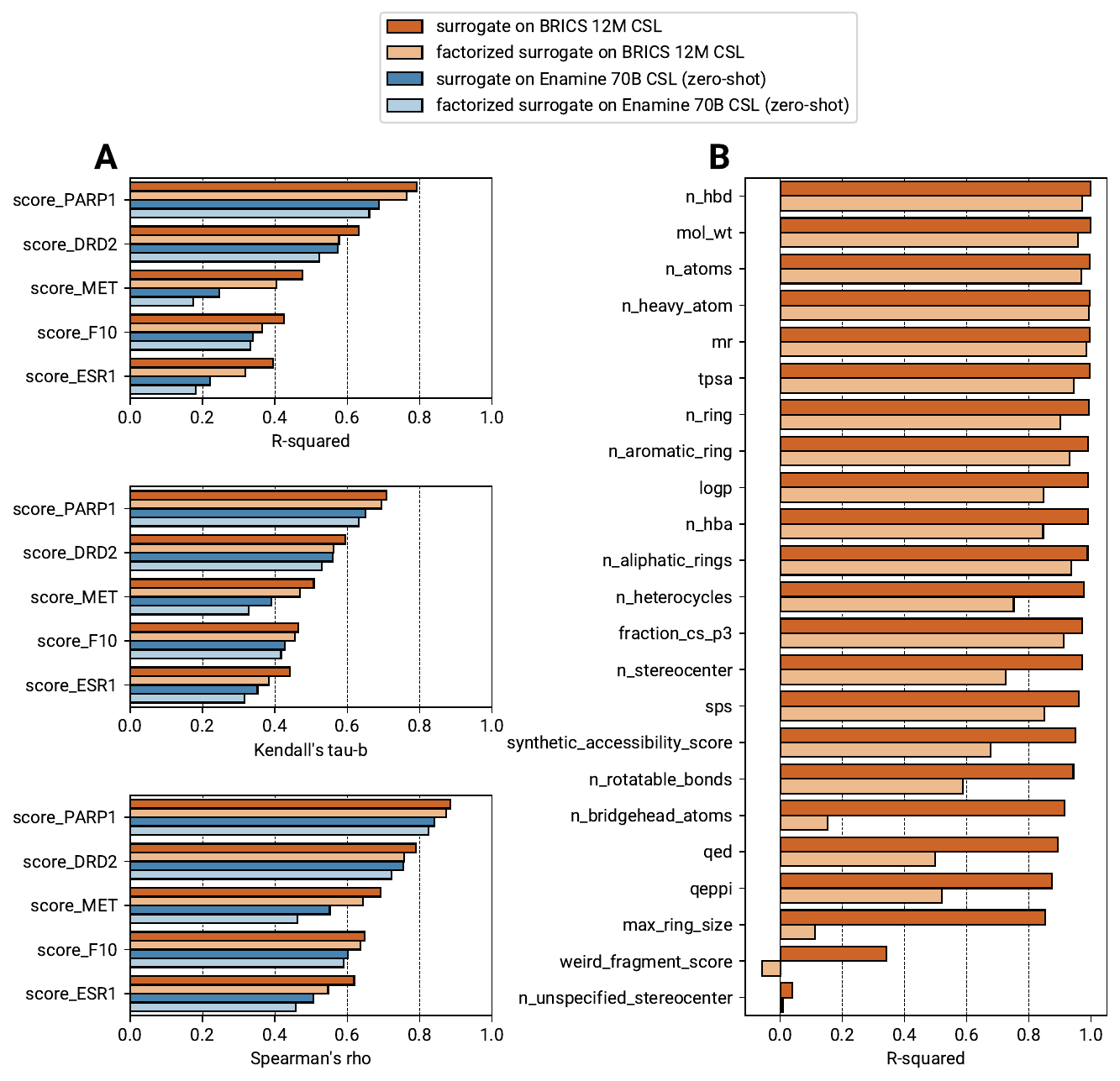}
  \caption{Accuracy of predicted (A) docking scores and (B) physicochemical properties for the original surrogate model as well as factorized version.}
  \label{fig:r2}
\end{figure}

Figure \ref{fig:r2} displays the R-squared for the surrogate (original and APEX-factorized) across all 28 endpoints considered in this study measured on a random sample of compounds from the 12M fully enumerated CSL. For the five docking score endpoints we additionally report rank correlation with Kendall's tau-b and Spearman's rho.

\begin{table}[ht]
\tiny
\centering
\caption{Description of all molecular properties and docking score endpoints used in this study.}
\label{tab:endpoints}
\begin{tabular}{ll}
\toprule
\textbf{Variable name} & \textbf{Description} \\
\midrule
\multicolumn{2}{l}{\textit{Docking scores}} \\
\texttt{score\_PARP1} & Docking score against PARP1 (poly ADP-ribose polymerase 1, an enzyme) \\
\texttt{score\_DRD2}  & Docking score against DRD2 (dopamine receptor D2, a GPCR) \\
\texttt{score\_MET}   & Docking score against MET (hepatocyte growth factor receptor, a kinase) \\
\texttt{score\_F10}   & Docking score against F10 (coagulation factor Xa, a protease) \\
\texttt{score\_ESR1}  & Docking score against ESR1 (estrogen receptor alpha, a nuclear receptor) \\
\midrule
\multicolumn{2}{l}{\textit{physicochemical properties}} \\
\texttt{mol\_wt}                         & Molecular weight (Da) \\
\texttt{logp}                            & Wildman--Crippen partition coefficient (lipophilicity) \\
\texttt{tpsa}                            & Topological polar surface area (\AA$^2$) \\
\texttt{n\_hbd}                          & Number of hydrogen bond donors \\
\texttt{n\_hba}                          & Number of hydrogen bond acceptors \\
\texttt{n\_rotatable\_bonds}             & Number of rotatable bonds \\
\texttt{n\_atoms}                        & Total number of atoms (including hydrogens) \\
\texttt{n\_heavy\_atom}                  & Number of heavy (non-hydrogen) atoms \\
\texttt{mr}                              & Wildman--Crippen molar refractivity \\
\texttt{n\_ring}                         & Total number of rings \\
\texttt{n\_aromatic\_ring}               & Number of aromatic rings \\
\texttt{n\_aliphatic\_rings}             & Number of aliphatic rings \\
\texttt{n\_heterocycles}                 & Number of heterocyclic rings \\
\texttt{n\_bridgehead\_atoms}            & Number of bridgehead atoms \\
\texttt{n\_stereocenter}                 & Number of specified stereocenters \\
\texttt{n\_unspecified\_stereocenter}    & Number of unspecified stereocenters \\
\texttt{fraction\_cs\_p3}               & Fraction of $sp^3$ carbon atoms (Fsp3) \\
\texttt{max\_ring\_size}                 & Size of the largest ring in the molecule \\
\texttt{qed}                             & Quantitative estimate of drug-likeness \citep{bickerton2012quantifying} \\
\texttt{qeppi}                           & Quantitative estimate of protein-protein interaction inhibitor drug-likeness \citep{kosugi2021quantitative} \\
\texttt{sps} & Spacial score, a topological indicator of small-molecule 3D complexity \citep{krzyzanowski2023sps} \\
\texttt{synthetic\_accessibility\_score} & Synthetic accessibility score (SA score) \citep{ertl2009estimation} \\
\texttt{weird\_fragment\_score}          & Score penalizing the presence of unusual or undesirable chemical fragments \\
\bottomrule
\end{tabular}
\end{table}

\subsection{GPU Implementation of Factorized Top-$k$ Search}\label{sec:gputopk}

The factorized top-$k$ search employed in APEX is particularly well suited for GPUs. Each operation (score calculation, element tracing, and index decoding) can be performed independently for each compound in the CSL. Moreover, NVIDIA's CCCL library \citep{cccl} provides an efficient batch-based AIR top-$k$ method \citep{zhang2023topk}, which we leverage in our implementation using a chain-of-batches strategy.

We first partition the CSL into batches of (reaction, first R-group assignment) pairs of some chunk size (e.g., one billion compounds) and evaluate scores, the two-dimensional pair $(\hat{\hat{c}}({\bf x}), \hat{\hat{f}}_0({\bf x}))$ denoting the APEX-predicted constraint violation and objective value, for all compounds in a batch on the GPU. For example, a batch can contain all compounds from the first three reactions (all R-groups fully enumerated) and all compounds from the fourth reaction where the first R-group assigned one of the first five eligible synthons, such that the total number of products is less than or equal to the specified chunk size.

Within a batch, compound scores are computed in parallel: CUDA blocks iterate over (reaction, first R-group assignment) pairs, while threads loop through subsequent R-group assignments. Synthon associative contributions are accumulated in shared memory for higher compute throughput. If a reaction has more than two R-groups, the remaining ones are processed with plain loops.

After score computation, results are passed to CCCL's AIR top-$k$ method to filter for the top-$k$ indices for that batch. For subsequent batches, previously selected elements are prepended to the score arrays before the next AIR top-$k$ call, and the indices within the full CSL are tracked, enabling chain-of-batches. We carefully trace the movement of elements: an index larger than $k$ means a new element from the current batch is within top $k$; otherwise the element is from previous batches but its location within top $k$ could have shifted. The kept indices array is updated accordingly.

Once the CSL is exhausted, each of the global top-$k$ indices within the library is decoded, again on GPU, using the (reaction, R-group assignment) mapping. The final results are then returned to the user for downstream processing (e.g., conversion of reaction and R-group assignment to SMILES).

\begin{table}
\centering
\small
\begin{tabularx}{\textwidth}{@{}l l l @{}}
\toprule
\textbf{Rule} & \textbf{Property} & \textbf{Value} \\
\midrule
\textbf{Lipinski Rule of 5} \citep{lipinski1997experimental} & Molecular weight & $\le 500$ Da \\
& logP & $\le 5$ \\
& H-bond donors & $\le 5$ \\
& H-bond acceptors & $\le 10$ \\
\midrule
\textbf{Veber} \citep{veber2002molecular} & Rotatable bonds & $\le 10$ \\
& TPSA & $\le 140$ \AA$^2$ \\
\midrule
\textbf{Pfizer 3/75} \citep{hughes2008physiochemical} & logP & $\le 3$ \\
& TPSA & $\ge 75$ \AA$^2$ \\
\midrule
\textbf{Wager CNS} \citep{wager2010moving} & Molecular weight & $\le 360$ Da \\
& logP & $\le 3$ \\
& TPSA & $\ge 40$ \AA$^2$, $\le 90$ \AA$^2$\\
& H-bond donors & $\le 1$ \\
\midrule
\textbf{Astex Rule of 3} \citep{congreve2003rule} & Molecular weight & $\le 300$ Da \\
& logP & $\le 3$ \\
& H-bond donors & $\le 3$ \\
& H-bond acceptors & $\le 3$ \\
& Rotatable bonds & $\le 3$ \\
& TPSA & $\le 60$ \AA$^2$\\
\bottomrule
\end{tabularx}
\caption{Constraint sets used for experiments in Figure \ref{fig:recall}.}
\label{tab:constraints}
\end{table}

\begin{figure}[ht]
  \centering
  \includegraphics[width=\textwidth]{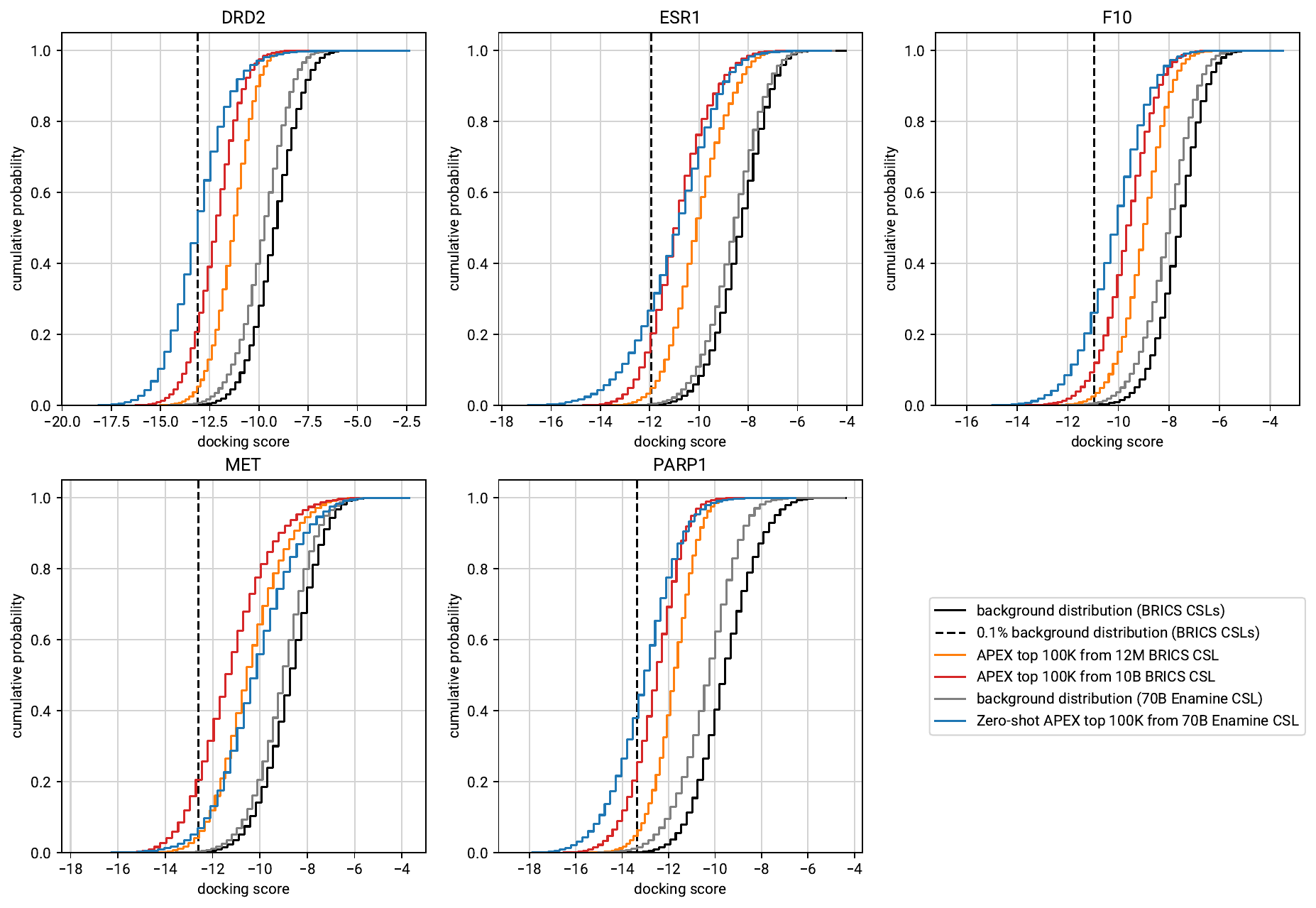}
  \caption{Docking scores for the APEX top-$k=\text{100,000}$ on the 10B library are enriched with respect to the background distribution and with respect to the top-$k$ set from the smaller 12M library. Lower scores are better (i.e., indicate better interaction between ligand and receptor).}
  \label{fig:scores}
\end{figure}

\subsection{Error Decomposition of the Factorized Surrogate}
\label{app:mse_decomposition}

For a given endpoint, the total prediction error of the APEX factorized surrogate, $\hat{\hat{f}}_i$, with respect to the ground truth oracle, $f_i$, can be decomposed into contributions from two distinct sources: the error of the surrogate $\hat{f}_i$ with respect to the oracle, and the error introduced by the factorizer in approximating the surrogate. The total error decomposes additively as:
\begin{equation}
    \hat{\hat{f}}_i(x) - f_i(x) = \underbrace{(\hat{f}_i(x) - f_i(x))}_{\text{surrogate error}} + \underbrace{(\hat{\hat{f}}_i(x) - \hat{f}_i(x))}_{\text{factorizer error}}.
    \label{eq:error_decomp}
\end{equation}
Squaring and taking expectations yields the following decomposition of the expected squared error:
\begin{align}
    \mathbb{E}\left[(\hat{\hat{f}}_i(x) - f_i(x))^2\right]
    &= \mathbb{E}\left[(\hat{f}_i(x) - f_i(x))^2\right] + \mathbb{E}\left[(\hat{\hat{f}}_i(x) - \hat{f}_i(x))^2\right] + 2\,\mathbb{E}\left[(\hat{f}_i(x) - f_i(x))(\hat{\hat{f}}_i(x) - \hat{f}_i(x))\right],
    \label{eq:mse_decomp}
\end{align}
where the first term is the surrogate MSE, the second term is the factorizer MSE, and the third term is a cross term capturing the correlation between the two error sources.

If the surrogate error and the factorizer error are approximately uncorrelated, i.e., if the cross term in  \eqref{eq:mse_decomp} is small in expectation, then the total MSE decomposes approximately as the sum of the two individual MSEs.

This decomposition is practically useful because each term can be reduced independently: the surrogate MSE is minimized by improving the molecular encoder $g_\theta$ or by training on more labeled data, while the factorizer MSE is minimized by improving the \texttt{ReactionFactorizer} $\hat{g}_\lambda$ or by training it on a larger or more diverse set of (reaction, R-group assignment) pairs --- crucially, without requiring any additional oracle labels.

\begin{figure}[ht]
  \centering
  \includegraphics[width=0.8\textwidth]{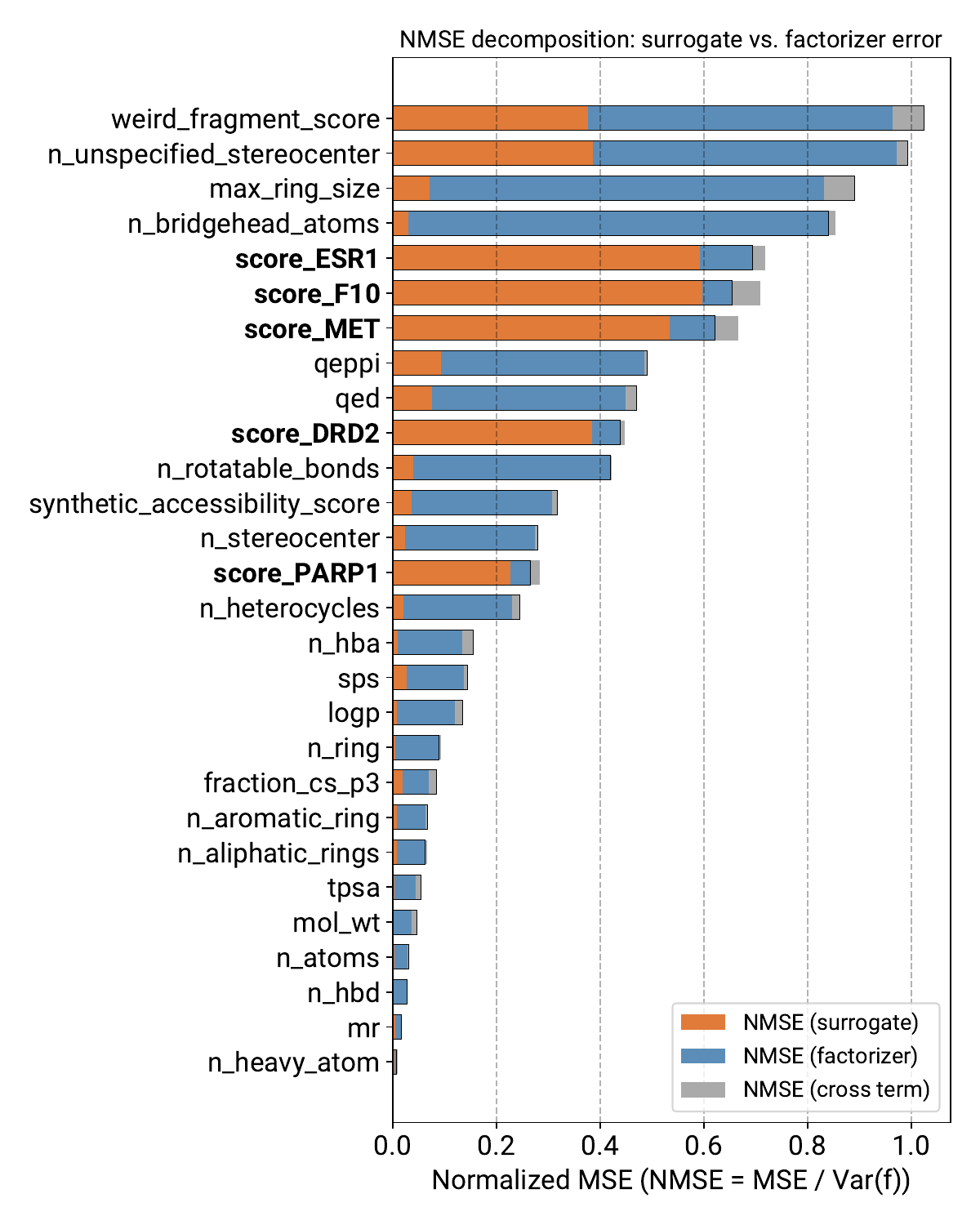}
\caption{Normalized mean squared error (NMSE $= \text{MSE} / \text{Var}(f)$) for the factorized surrogate $\hat{\hat{f}}_i$ with respect to the ground truth oracle $f_i$, decomposed into three additive components per endpoint as derived in Appendix \ref{app:mse_decomposition}: surrogate error (orange), factorizer error (blue), and cross term (gray). The three components sum to the total NMSE for each endpoint, indicated by the black outline enclosing all three components. The cross term is negligible for most endpoints, indicating that the two error sources are largely independent and that the total NMSE is well approximated by the sum of the surrogate and factorizer MSEs alone.}
\label{fig:nmse}
\end{figure}

Figure \ref{fig:nmse} plots the decomposition of the normalized MSE across all 28 endpoints. We note that the cross term is negligible across all endpoints, confirming that the surrogate and factorizer errors are largely independent; this means that improvements to the surrogate encoder and the factorizer can be pursued separately to reduce the total NMSE. The relative contributions of the two error sources vary across endpoints: for some, such as the docking scores, the surrogate error (orange) dominates, while for others, the factorizer error (blue) is the primary contributor. This can serve as a guide for where architectural improvements are likely to be most impactful.

\subsection{Constraint Sets}\label{constraints}

Table \ref{tab:constraints} provides details on the constraints used in this paper's experiments. We note that all of the computational endpoints and properties described in Table \ref{tab:endpoints} can be used as objectives or constraints in an APEX query for the model trained as part of this paper.

\subsection{Architecture and Training Details}
\label{app:architecture}

Below we provide a description of the model architectures and key hyperparameters  used in this paper; the overall setup is intentionally lightweight and we did not perform extensive hyperparameter tuning, instead opting for sensible defaults. We refer the reader to the code repository (\url{https://github.com/NumerionLabs/apex}) for full implementation details.

\textbf{Message-passing neural network (MPNN).}
Both the surrogate encoder and the factorizer's synthon encoder are built on the same underlying MPNN architecture (implemented in \texttt{apex/nn/mpnn.py}), which operates on 2D molecular graphs.
Each atom is featurized using a standard set of atomic features (atom type, degree, formal charge, chirality, aromaticity, etc.) and each bond is featurized using bond-level features (bond type, ring membership, stereochemistry, etc.).
These are projected into a $d_\text{node}$-dimensional node embedding and $d_\text{edge}$-dimensional edge embedding by learned linear layers.
The network then applies $L$ rounds of edge-conditioned message passing, in which each node aggregates messages from its neighbors, where each message is conditioned on the corresponding edge embedding.
A graph-level representation is obtained by summing the final node embeddings and applying a final linear transformation to produce a $d_\text{graph}$-dimensional output.
In all models used in this paper, all hidden dimensions are set to $d=64$ and the number of message-passing layers is $L = 4$.

\textbf{Surrogate.}
The surrogate model $g_\theta$ is a \texttt{LigandEncoder} (implemented in \texttt{apex/nn/encoder.py}) consisting of the MPNN described above, followed by a linear projection to the $d = 64$-dimensional embedding space.
Task-specific predictions are obtained via a shared \texttt{LinearProbe} consisting of $28$ task-specific linear heads (one per endpoint). The total parameter count for the surrogate encoder is 159,168.
The probe adds $28 \times 64 = 1{,}792$ weight parameters and 28 bias parameters, for an additional 1,820 parameters.

\textbf{Factorizer.}
The \texttt{APEXFactorizer} (implemented in \texttt{apex/nn/apex.py}) mirrors the CSL hierarchy with a corresponding stack of encoders.
At the bottom, a \texttt{SynthonGraphEncoder} --- an instance of the same \texttt{LigandEncoder} MPNN architecture as the surrogate encoder --- produces a $d_\text{hidden} = 64$-dimensional embedding for each synthon from its molecular graph.
R-group embeddings are produced by a deep set network consisting of two \texttt{ResNetMLP} submodules with mean pooling applied between them over the synthons belonging to each R-group.
Reaction embeddings are produced analogously by two further \texttt{ResNetMLP} submodules with sum pooling over the R-groups belonging to each reaction.
Finally, a \texttt{RgroupAssociativeKeyEncoder} (a \texttt{ResNetMLP}) maps each (R-group, reaction) pair to a $d \times d_U = 64 \times 64$ key matrix, and a \texttt{SynthonAssociativeValueEncoder} (a \texttt{ResNetMLP}) maps each synthon to a $d_U = 64$-dimensional value vector; their product gives the $d = 64$-dimensional synthon associative embedding $u_{r,s}$, as per \eqref{assoc_embeds}.
Each \texttt{ResNetMLP} uses $d_\text{hidden} = 64$, $n_\text{layers} = 4$, and layer normalization.
The total parameter count for the factorizer is 676,992.

\textbf{Training.}
Training hyperparameters are specified in \texttt{configs/apex.yaml} in the associated code repository.
Both models were trained using the Adam optimizer with learning rate $10^{-3}$ for up to 100,000 iterations.
The surrogate was trained on 80\% of the 1M compound BRICS-based CSL (with the remaining 20\% held out for validation). The factorizer was trained on the 12M compound BRICS-based CSL in the absence of labels, minimizing the reconstruction error of the frozen surrogate embeddings \eqref{factorizer_opt}.
For the 10B BRICS CSL, the factorizer was additionally fine-tuned on compounds from the library (using the same embedding reconstruction task).

\subsection{Comparison with Thompson Sampling}

\begin{figure}[ht]
  \centering
  \includegraphics[width=0.5\textwidth]{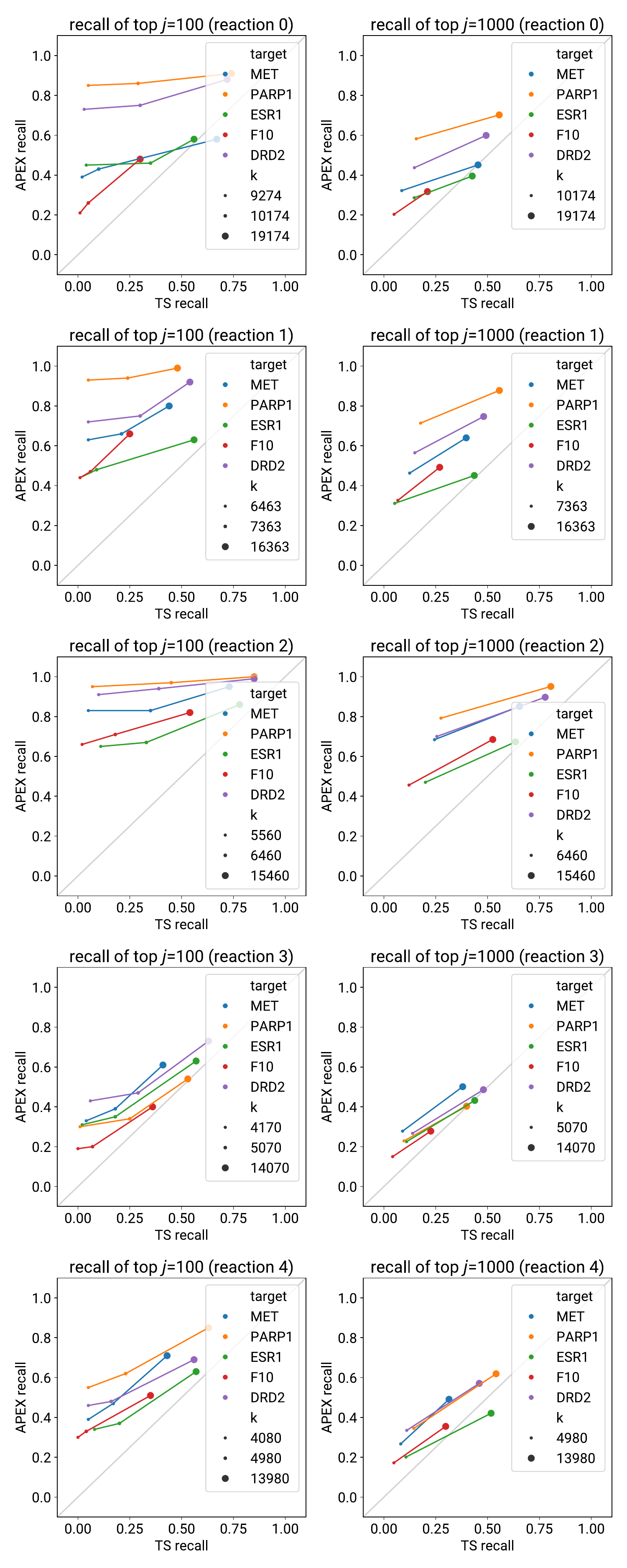}
  \caption{Top-$j$ recall for APEX and Thompson sampling (TS) using matched evaluation budgets. APEX search run using $k$ set to the number of total evaluations for TS. Thompson sampling comparison was run using three and ten warmup steps for two- and three-component reactions, respectively, and 10, 1000, or 10,000 iterations of Thompson sampling.}
  \label{fig:ts}
\end{figure}

Figure \ref{fig:ts} plots the recall of APEX against Thompson sampling across the five most prevalent reactions in the 12M compound CSL and against the five targets considered in this paper.

\subsection{Recall of Top-$j$ Compounds at Increasing Evaluation Budget}
\begin{figure}[ht]
  \centering
  \includegraphics[width=\textwidth]{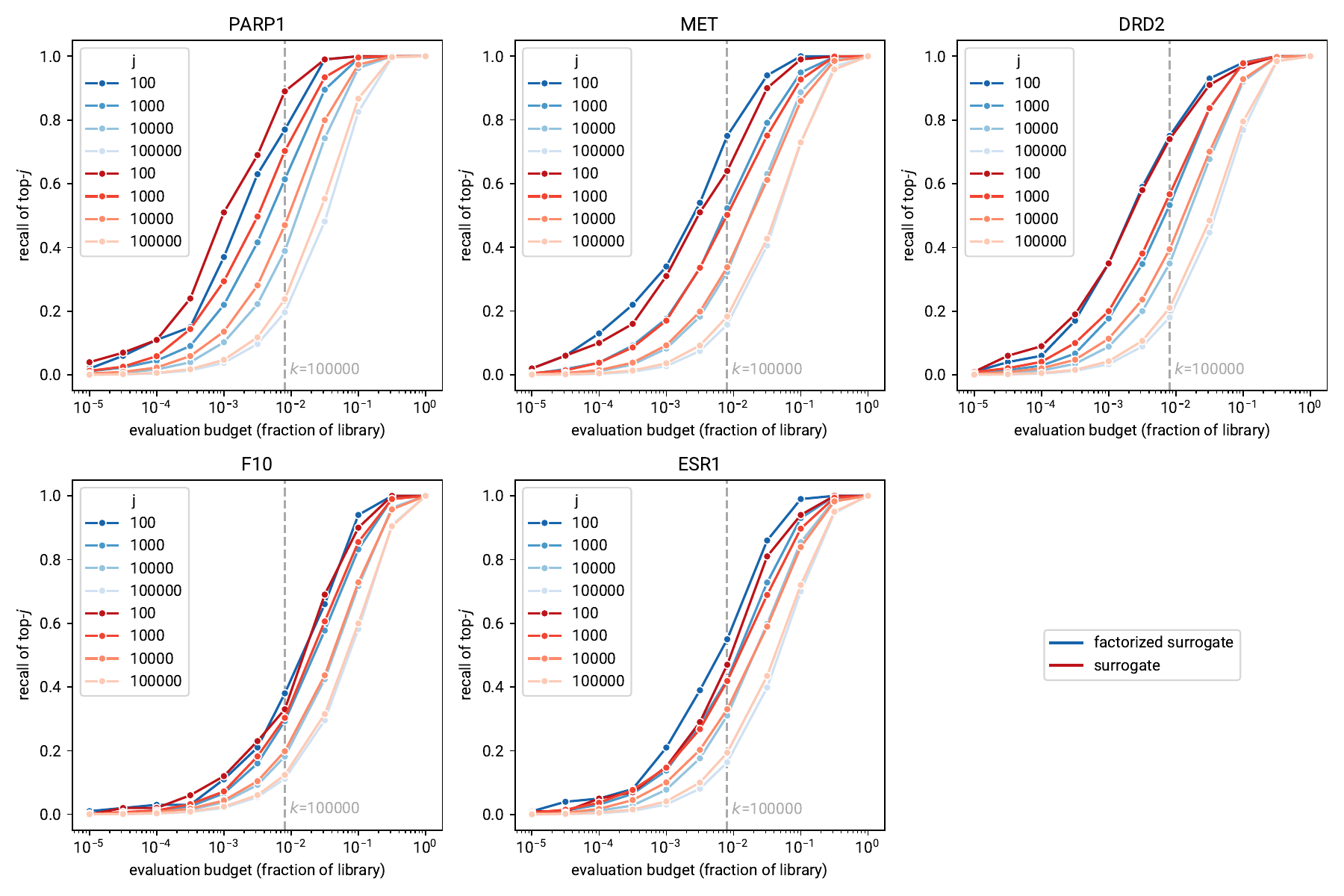}
  \caption{Recall of ground truth top-$j$ compounds by the surrogate model (red) and factorized surrogate (blue) at different evaluation budgets. No constraints were imposed. Dashed line corresponds to a budget of $k=\text{100,000}$ compounds, which was used for the evaluations in Figure \ref{fig:recall}.}
  \label{fig:recall_at_k}
\end{figure}

Figure \ref{fig:recall_at_k} shows the recall of top compounds (in the absence of constraints) as a function of increasing evaluation budget, expressed as the fraction of the library evaluated with the oracle.

\subsection{Score-Based Constraints and Composite Objectives}

\begin{figure}[ht]
  \centering
  \includegraphics[width=\textwidth]{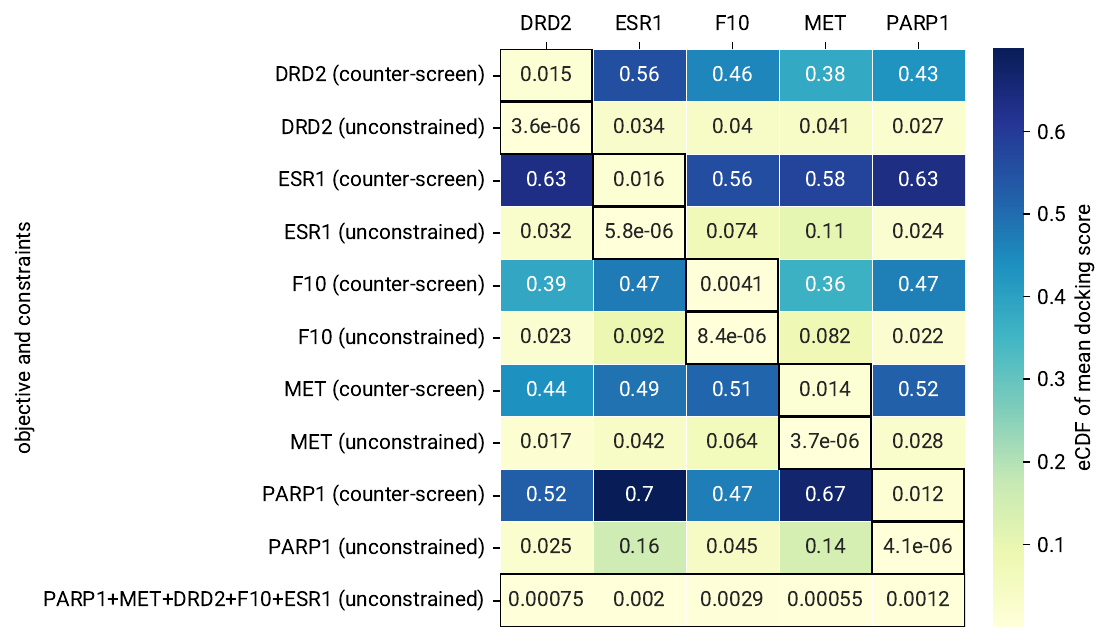}
  \caption{Inclusion of constraints on non-objective docking scores allows for APEX to be used in a counter-screening fashion. Each row is the result of a single APEX search ($k=\text{100,000}$), either unconstrained or with ``counter-screening'' constraints (non-objective docking scores $>50$th percentile). Cells are outlined if they were used as the objective, and values are the eCDF of the mean docking score for the top 100 molecules after re-ranking by the true objective. (\textit{Last row)} APEX search with composite objective of all five targets' docking scores.}
  \label{fig:counterscreen}
\end{figure}

To further test the robustness of the surrogate docking score predictions, we ran APEX search in a counter-screening scenario, where one target is chosen as the objective to minimize and constraints are added that the other four targets all score above their 50th percentile. Figure \ref{fig:counterscreen} shows the mean docking scores of the best 100 compounds (re-ranked by the true objective) after a $k=100,000$ search (BRICS 12M library), represented in terms of their eCDF. While these counter-screening constraints are generally effective at increasing the ``selectivity'' of the top compounds, they do result in worse absolute docking scores for the objective. We also tested defining a composite objective as the sum of all five targets' docking scores, which proved quite effective at finding compounds that score well across all targets.

\end{document}